\documentclass[review]{IEEEtran}

\usepackage{amsfonts}
\usepackage{amsmath}
\usepackage{amsmath,epsfig,amsmath}
\usepackage{amsfonts,amssymb,amsthm}
\usepackage{epsfig}
\usepackage{float}
\usepackage{graphicx}
\usepackage{color,hyperref}
\usepackage{multirow}
\usepackage{tipa}
\usepackage{picins}
\usepackage{color}
\usepackage{booktabs}
\usepackage{balance}

\begin{document}




\title{Heterogeneous Trajectory Forecasting \\via Risk and Scene Graph Learning}

\author{Jianwu Fang, Chen Zhu, Pu Zhang, Hongkai Yu, and Jianru Xue
\thanks{
J. Fang and C. Zhu are with Chang'an University, Xi'an, China, and Jianwu Fang is also a visiting scholar at NExT++ Centre, School of Computing, National University of Singapore, Singapore
        {(fangjianwu@chd.edu.cn)}.}%
\thanks{P. Zhang is with the KargoBot, Beijing, China
        {(zhangpu94@gmail.com)}.}%
\thanks{J. Xue is with the Institute of Artificial Intelligence and Robotics, Xi'an Jiaotong University, Xi'an, China
        {(jrxue@mail.xjtu.edu.cn)}.}%
   \thanks{H. Yu is with the Department of Electrical Engineering and Computer Science, Cleveland State University, Cleveland, USA
        {(h.yu19@csuohio.edu)}.}
}

\maketitle

\begin{abstract}
Heterogeneous trajectory forecasting is critical for intelligent transportation systems, but it is challenging because of the difficulty of modeling the complex interaction relations among the heterogeneous road agents as well as their agent-environment constraints.  In this work, we propose a risk and scene graph learning method for trajectory forecasting of heterogeneous road agents, which consists of a Heterogeneous Risk Graph (HRG) and a Hierarchical Scene Graph (HSG) from the aspects of agent category and their movable semantic regions. HRG groups each kind of road agent and calculates their interaction adjacency matrix based on an effective collision risk metric. HSG of the driving scene is modeled by inferring the relationship between road agents and road semantic layout aligned by the \emph{road scene grammar}. Based on this formulation, we can obtain effective trajectory forecasting in driving situations, and comparable performance to other state-of-the-art approaches is presented by extensive experiments on the nuScenes, ApolloScape, and Argoverse datasets.
\end{abstract}



\begin{IEEEkeywords}
Heterogeneous trajectory forecasting, intelligent vehicles, collision risk, road scene graph, MLP
\end{IEEEkeywords}
\markboth{LATEX}%
{}


\section{Introduction}
\label{section1}
\IEEEPARstart{F}{or} safe driving, more and more efforts in autonomous driving or assisted driving systems are being devoted to predicting the trajectories of surrounding road participants. It is at the forefront of accurate action and planning decisions for intelligent driving systems, to minimize collision risks and maximize traffic capacity. However, the open, dynamic, and complex driving scene still makes it difficult to accurately forecast the trajectories of the road participants because of the heavy uncertainty of the future and the inadequate characterization of traffic rules. Many road accidents are caused by the wrong estimation of the future state of the traffic scene and the untimely avoidance of collision \cite{DBLP:journals/tits/LinCCS22,fang2022traffic,DBLP:journals/corr/abs-2212-09381}.

 \begin{figure}[!t]
  \centering
 \includegraphics[width=\hsize]{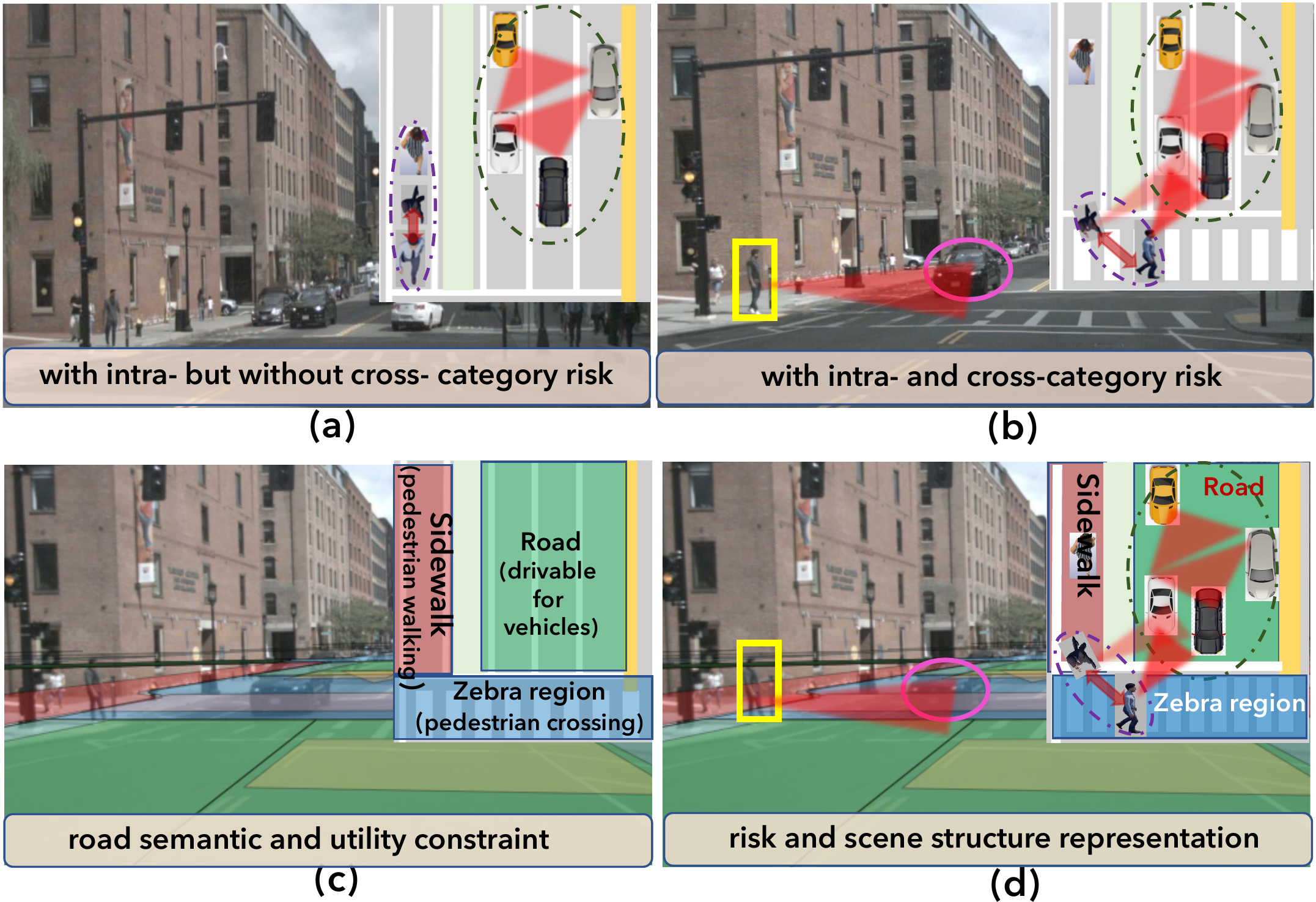}
  \caption{The illustration for risk reasoning and road topological and utility constraint (drivable, crossing, walking, etc.) for heterogeneous road participants in a mixed traffic scene. The red triangles mean the inter-risk of different road agents concerning differing road semantic regions.}
  \label{fig1}
  \vspace{-1em}
\end{figure}

In recent years, most of the trajectory forecasting studies are data-driven approaches with various deep learning models \cite{DBLP:conf/cvpr/AlahiGRRLS16,DBLP:conf/cvpr/GuptaJFSA18,DBLP:conf/cvpr/ZhangO0XZ19,su2022trajectory,DBLP:journals/tits/ChenWHSLY23} and are modeled by taking several historical frames of participant locations as the observation and several future frames of points to be predicted. 
In existing works, the interaction model between road agents is the key component for interpreting their social-related behaviors with the investigation of physical dynamics (e.g., social force \cite{DBLP:conf/cvpr/MehranOS09}) and psychological knowledge (e.g., moving intentions \cite{DBLP:conf/iccv/GiraseGM0KMC21}) in road society. Over one decade, a large part of the trajectory prediction works concentrate on the interaction model among single kind of participants, such as pedestrians or vehicles, and various interaction models are inspired by the social-force model with a physical or geometric relation, such as Social-LSTM \cite{DBLP:conf/cvpr/AlahiGRRLS16}, SR-LSTM \cite{DBLP:conf/cvpr/ZhangO0XZ19}, Social-Attention \cite{DBLP:conf/icra/VemulaMO18}, and so on. However, the road scene is a highly socialized environment, where many kinds of participants (pedestrians, vehicles, cyclists, etc.) commonly exist at the same time except in highway scenarios. Meanwhile, heterogeneous road participants often take turns to occupy the road right \cite{DBLP:journals/corr/abs-2211-00385}. Their interaction forms the basis of the ``\emph{theory of mind}" for road participants to assess the unknown intention of other targets \cite{DBLP:series/leus/Restivo07}.

Some works are aware of heterogeneous trajectory forecasting and formulate a few interaction models to infer the relationships among different kinds of road participants \cite{DBLP:conf/iccv/Zheng0ZTN0021,DBLP:journals/corr/abs-2111-14973,DBLP:conf/cvpr/KimPLKKKKC21,DBLP:journals/ral/WangWXC22,DBLP:conf/icra/GillesSTSM22,DBLP:conf/iclr/GirgisGCWDKHP22}, such as Trajectron++ \cite{DBLP:conf/eccv/SalzmannICP20}, LaPred \cite{DBLP:conf/cvpr/KimPLKKKKC21}, GOHOME \cite{DBLP:conf/icra/GillesSTSM22}, and so on. These heterogeneous trajectory forecasting works on the road scene focus on the interaction between different kinds of road participants, and commonly encode the scene context information by feeding the semantic segmentation images into some convolution neural networks. This kind of consideration for road semantic information omits the natural constraint of traffic rules and the moving risk of participants when competing for the road right.

\textbf{Motivations:} Considering this, this work begins the formulation from two motivations. (1) Different kinds of road participants (e.g., pedestrians or vehicles) have a differing extent of collision risk which should be adaptively activated or eliminated with the constraint of road semantic types. As shown in Fig. \ref{fig1} (a), if the pedestrian and the vehicle move in the sidewalk and road region separately. There is no risk between them, while the between-category risk of them should be activated if they compete the road right, as shown in Fig. \ref{fig1}(b). (2) The structural-semantic relation of the road scene is important and the structural relation should obey the natural topology and utility constraint. For example, the road segments naturally link the drivable area and the zebra region, which is an important prior for the future movements of vehicles and pedestrians, as shown in Fig. \ref{fig1}(c). Traditional approaches take the road semantic images into the deep learning model, which is without the explicit consideration of the natural topology and the utility constraint. With the road semantic layout and the constraint of the participants' movable area, collision risk can be comprehensively analyzed (Fig. \ref{fig1}(d)).

Therefore, we propose a risk and scene graph learning model which towards trajectory forecasting for heterogeneous traffic participants. In particular, the movement risks, road scene graph, and the relation between risk and road semantic layout are considered together. This formulation consists of a Heterogeneous Risk Graph (HRG, in Sec. \ref{cmsrg}), a Hierarchical Scene Graph (HSG, in Sec. \ref{rssg}), and a graph embedding fusion-based trajectory forecaster (Sec. \ref{graphfusion}). HRG groups each kind of road participant and calculates their interaction adjacency matrix based on a self-defined collision risk metric. HSG is modeled by inferring the road participant attributes and their topological relationship aligned with the \emph{road scene grammar} \cite{DBLP:conf/eccv/DevaranjanKF20}. Graph embedding fusion-based trajectory forecaster fuses the embedding of HRG and HSG to spatial-temporally learn the core graph correlation in trajectory forecasting. Extensive experiments are conducted with the nuScenes, ApolloScape, and Argoverse datasets, and superior performance to other state-of-the-art methods is obtained. To summarize, the \textbf{contributions} of this work are:

\begin{itemize}
\item Different from previous works, we explicitly model the influence of road semantic layout aligned with the road scene grammar to fulfill the trajectory forecasting of heterogeneous agents.
\item We model the risk and scene graph of heterogeneous participants by learnable Heterogeneous Risk Graph and Hierarchical Scene Graph, where the natural topology and utility constraint of the road scene is modeled.
\item The proposed method is explainable and comparable to other state-of-the-art methods by extensive evaluations on three challenging datasets, \emph{i.e.}, nuScenes \cite{DBLP:conf/cvpr/CaesarBLVLXKPBB20}, ApolloScape \cite{DBLP:journals/pami/HuangWCZGY20}, and Argoverse \cite{DBLP:conf/cvpr/ChangLSSBHW0LRH19} datasets.
\end{itemize}

The remainder of this work is organized as follows. Sec. \ref{relatework} briefly reviews the precedent works. The proposed method is presented in Sec. \ref{method}. Sec. \ref{expe} evaluates the proposed method and the comparison analysis with other methods, and the conclusion is given in Sec. \ref{con}.
\begin{figure*}[!t]
  \centering
 \includegraphics[width=\hsize]{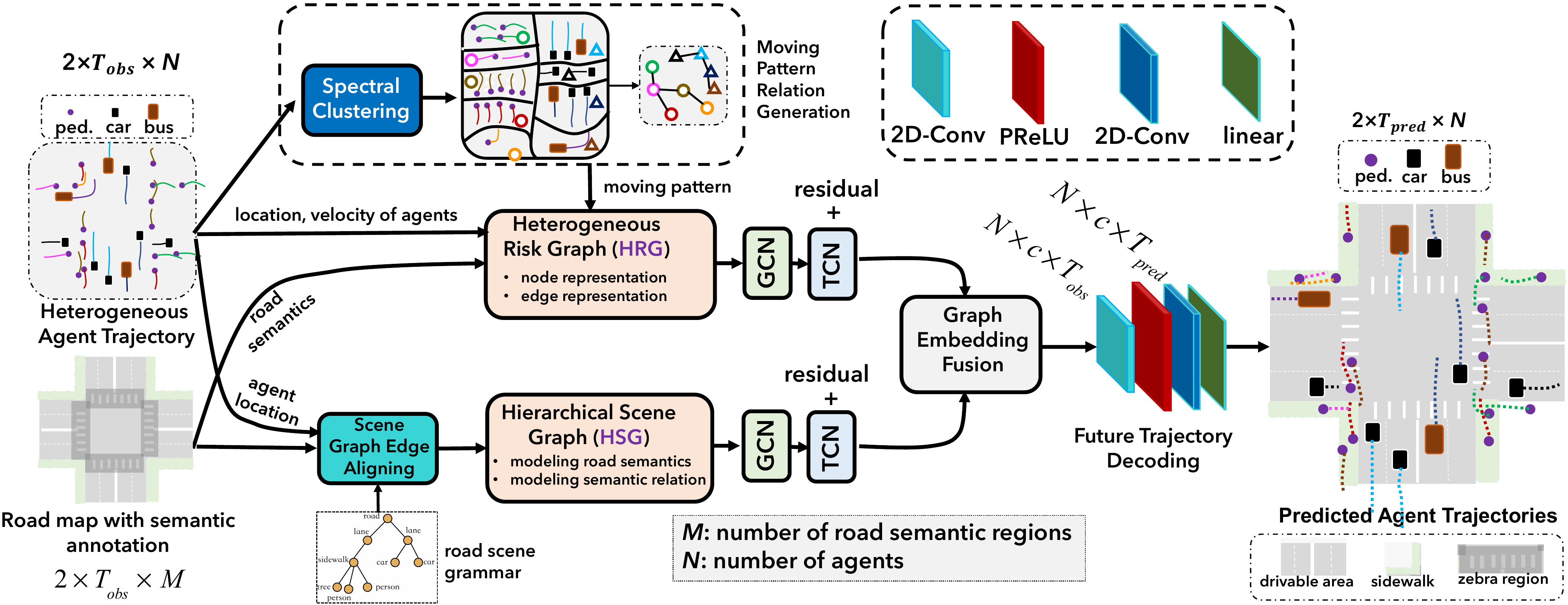}
  \caption{The main architecture of the proposed method. With the input of heterogeneous agent trajectories and the road map with semantic annotation, they are mutually fused and encoded by the  \textcolor{cyan}{Heterogeneous Risk Graph (HRG)} and \textcolor{magenta}{Hierarchical Scene Graph (HSG)}. Among them, we investigate the moving pattern relation of different groups of agents by a spectral clustering algorithm \cite{DBLP:conf/nips/NgJW01}. After that, HRG encodes the relations of the locations, velocity, the moving pattern of agents, and road semantic layout to fulfill an explainable risk reasoning formulation. HSG encodes the semantic linking structure of agents concerning the road scene with \emph{road scene grammar} by the Scene Graph Edge Aligning module (to be described in Sec. III-C). HRG and HSG are followed by a conjunction of Graph Convolution Network (GCN) and Temporal Convolution Network (TCN) for inferring the spatial-temporal scene evolution, and then fuse to obtain a risk and scene graph-aware trajectory embedding. Finally, the fused graph embedding is decoded by two 2D convolution layers, one Parametric Rectified Linear Unit (PReLU) \cite{DBLP:conf/iccv/HeZRS15}, and one linear layer for future trajectory generation, where the dimension of the convolution is noted and the last value $c$ is the number of the parameters in the trajectory regression model (to be described in the loss function).}
  \label{fig2}
\end{figure*}
\section{Precedent Work}
\label{relatework}
Previous trajectory prediction methods focus on the homogeneous interaction model among a single kind of participants and the relation representation between heterogeneous participants. In addition, some trajectory prediction works involve risk reasoning for safety awareness and are briefly discussed.

\subsection{Homogeneous Interaction in Trajectory Prediction}
The pioneering interaction model in trajectory prediction can be the Social-Force model in Social-LSTM \cite{DBLP:conf/cvpr/AlahiGRRLS16}, which converts the influencing factors between targets and obstacles into corresponding gravitational and repulsive forces \cite{karamouzas2014universal} and uses these forces and mathematical analytical formulas to build a model to infer the target movement path. Inspired by that, many excellent social state embedding models, such as Social-LSTM \cite{DBLP:conf/cvpr/AlahiGRRLS16}, Social-GAN \cite{DBLP:conf/cvpr/GuptaJFSA18}, SR-LSTM \cite{DBLP:conf/cvpr/ZhangO0XZ19}, Social-Attention \cite{DBLP:conf/icra/VemulaMO18}, and Social-BiGAT \cite{DBLP:conf/nips/KosarajuSM0RS19} are proposed and generate good performance. Within this field, many trajectory forecasting works focus on Bird's Eye View (BEV) scenes captured by the UAV platforms from campus or highway occasions \cite{DBLP:conf/cvpr/XuW0022}. 

As for the interaction model in trajectory prediction, \emph{social state-level graph relation} and \emph{feature-level attention} between the agents are considered in the social and time dimensions. The social state-level relation is commonly modeled by the physical distance, directions, and velocity between targets, such as SGSG \cite{DBLP:journals/corr/abs-2010-05507} and EvolveGraph \cite{DBLP:conf/nips/LiYTC20}. For example, EvolveGraph \cite{DBLP:conf/nips/LiYTC20} explores explicit relation structure recognition and multi-modal prediction by latent interaction graphs among multiple agents. The work \cite{DBLP:conf/iros/MerschHZSR21} encodes the past interaction states between vehicles and combines the correlation of intention and interaction in the space and time dimensions. Therefore, spatial-temporal interaction models (e.g., GCN+GRU \cite{DBLP:journals/corr/abs-2109-12764}) take a large role in vehicle trajectory prediction \cite{DBLP:journals/corr/abs-2109-12764,DBLP:journals/itsm/LinLBQ22}.

For the feature-level attention model in the interaction, self-attention models are the typical choice  \cite{DBLP:conf/icpr/GiuliariHCG20,DBLP:journals/corr/abs-2106-08417}. For example, the work \cite{DBLP:conf/nips/KamraZT0020} proposes Fuzzy Query Attention (FQA) which generates the keys, queries, and responses from all agent pairs (\emph{i.e.}, senders and receivers) and combines the responses by fuzzy decisions. AgentFormer \cite{DBLP:conf/iccv/0007WOK21} models the effects of agents in the social and time dimension simultaneously, which simulates the potential intentions of all agents and produces a socially reasonable future trajectory. 

In a nutshell, these interaction models in trajectory prediction present excellent forecasting performance for homogeneous participants, while the road scene is a highly socialized and mixed environment and many kinds of road participants may appear at the same time. The interaction between different types of road participants is preferred for driving scenes.

\subsection{Heterogeneous Trajectory Forecasting}

Heterogeneous trajectory forecasting generates corresponding future trajectories for different kinds of road participants. Recently, thanks to the release of large-scale datasets for driving scenes, such as nuScenes \cite{DBLP:conf/cvpr/CaesarBLVLXKPBB20}, Waymo \cite{DBLP:conf/cvpr/SunKDCPTGZCCVHN20}, and Euro-PVI \cite{DBLP:conf/cvpr/BhattacharyyaRF21}, rich and diverse interactive information between heterogeneous participants becomes feasible. 

Heterogeneous trajectory prediction involves the \emph{heterogeneous interaction modeling} \cite{DBLP:conf/iccv/HuangBLMW19,DBLP:conf/cvpr/MohamedQEC20,DBLP:conf/cvpr/TangNHSZ20} and the \emph{future hypothesis estimation} \cite{DBLP:journals/corr/abs-2111-14973,DBLP:conf/eccv/ParkLSBKFJLM20,DBLP:conf/iccv/BiFMWD19} constrained by the road layout. Trajectron ++ \cite{DBLP:conf/eccv/SalzmannICP20} is one typical heterogeneous interaction work that builds a graph-structured cyclic model which combines agent dynamics of different road participants and the semantic road map. TrafficPredict \cite{DBLP:conf/aaai/MaZZYWM19} constructs a 4D graph to infer the interaction over the spatial-temporal-instance dimension and instance category dimension, which has an assumption that the instances in the same category could have similar dynamics or interaction patterns. TraPHic \cite{DBLP:conf/cvpr/ChandraBBM19} considers the weighted horizon and heterogeneous interaction to prioritize the importance of the different kinds of road agents with conflicting moving directions. Unlimited Neighborhood Interaction (NLNI) \cite{DBLP:conf/iccv/Zheng0ZTN0021} models the interactions between dynamic agents from the spatial-temporal-category and their unlimited neighborhood, which enforces a wider correlation with the surrounding agents regardless of their numbers of them and achieves a heterogeneous graph over different kinds of agents. 

Future hypothesis estimation predicts the probability of future trajectories and aims to rule out the invalid trajectories. For example, MultiPath \cite{DBLP:conf/corl/ChaiSBA19} and MultiPath++ \cite{DBLP:journals/corr/abs-2111-14973} estimate the probability of future trajectories and reduce the probability of unlikely trajectories. This formulation is also involved in the work \cite{zhu2022motion} by unlikelihood training in continuous space. In addition, some works introduce the kinematics of pedestrians and vehicles to optimize the prediction. For example, JPKT \cite{DBLP:conf/iccv/BiFMWD19} adds the orientation variable in multi-variable Gaussian distribution in future vehicle trajectories. DATF \cite{DBLP:conf/eccv/ParkLSBKFJLM20} models the environment's scene context and the interaction between multiple surrounding agents for penalizing implausible trajectories that either conflict with the other agents or are outside of valid drivable areas. DATF closes to our work while we are based on the point of risk and scene semantic context learning that explicitly models the heterogeneous risk graph and hierarchical scene graph aligned with road scene grammar.

\subsection{Risk Reasoning in Trajectory Forecasting}
In complex driving situations, collision avoidance is the crucial component of system safety, which requires a comprehensive analysis of the collision risk. Sense Learn Reason Predict (SLRP) framework \cite{DBLP:conf/icra/0001ZZWZ21} utilizes a planning-based method to fulfill collision-free trajectory prediction in a synthetic dataset. Risk has a direct relation to the prediction uncertainty, such as the interaction uncertainty \cite{DBLP:journals/corr/abs-2110-13947,zhang2023towards}  and class uncertainty \cite{ivanovic2022heterogeneous}. Collaborative Uncertainty \cite{DBLP:journals/corr/abs-2110-13947} is an interaction uncertainty for road participants, which can assist the driving system to obtain a better multi-agent prediction trajectory. Joint-$\beta$-Conditional Variational Autoencoder (Joint-$\beta$-cVAE) \cite{DBLP:conf/cvpr/BhattacharyyaRF21} simulates shared potential risk space to capture the impact of pedestrian interaction on future trajectory distribution. Recently, Wang \emph{et al.} \cite{DBLP:journals/tits/WangAW22} leverage the trajectory prediction to calculate the driving risk by estimating three factors of lane-change maneuver probability, collision probability, and expected crash severity, where better crash prediction ability is obtained. However, these works in trajectory forecasting ignore the road semantic constraint in explicit mode.

\section{Our Approach}
\label{method}
\subsection{Problem Formulation}

The trajectory forecasting problem is formally defined as a regression problem with several historical frames of target locations as the observation and some future frames of target locations to be predicted and defined as
\begin{equation}
\mathcal{X}_{pred}=\mathcal{P}(\mathcal{X}_{obs}, \mathcal{S}_{obs}),
\label{eq:1}
\end{equation}
where $\mathcal{X}_{obs}$ is the observation and commonly constructed by $\{{\bf{p}}^i_t\}_{t=1}^{T_{obs}}$, and $\mathcal{X}_{pred}$ is the set of future target locations $ \{{\bf{p}}^i_t\}_{t={T_{obs}+1}}^{T_{obs}+T_{pred}}$, where ${\bf{p}}=(x,y)$ is the  2D target coordinates at each time step, $T_{pred}$ is the prediction window and $i \in $[1,$N$] is the target index of participant category (e.g., pedestrians, cars, cyclist, etc.). Notably, this work introduces the road semantic information $\mathcal{S}_{obs}$ which is denoted as $\{S_1,...,S_{T_{obs}}\}$, where $S$ specifies the road semantic layout at each frame. Consequently, the main problem of this work is to model the regression model $\mathcal{P}(.,.)$ with the input of target locations and road semantic layout.

We all know that the movement safety of each road participant is influenced by its surrounding objects and the road semantic constraint. Meanwhile, different kinds of road participants at distinct road semantic regions have distinct safety degrees. This fact implies a category and semantic aware risk relation within the road participants, and an agent (road participant)-semantic scene constraint in the trajectory forecasting problem. Therefore, when modeling the future trajectory regression model  $\mathcal{P}(.,.)$, we approach it from the learning of a \emph{Heterogeneous Risk Graph (HRG)} $\mathcal{G}_{HRG}$ and a \emph{Hierarchical Scene Graph (HSG)} $\mathcal{G}_{HSG}$, which are then encoded by the spatial-temporal graph learning in prediction. Fig. \ref{fig2} presents the method architecture of this work.

\subsection{Heterogeneous Risk Graph (HRG)}
\label{cmsrg}

Heterogeneous Risk Graph defines the interaction relation within different agents from the consideration of collision risk, which is different from recent works that focus on the interaction of agents themselves from physical or geometrical aspects \cite{DBLP:conf/iccv/Zheng0ZTN0021,DBLP:conf/eccv/SalzmannICP20}. The collision risk metric prefers an insight from the road utility constraint (drivable, crossing, walking, etc.) and the movable pattern difference of the agents.

Formally, assume the heterogeneous risk graph at time $t$ be $\mathcal{G}^{HRG}_t=({\bf{A}}_t,{\bf{E}}_{t}^{HRG}, {S}_t)$. For $N$ agents in the observation, ${\bf{A}}_t =\{{\bf{a}}_t^1,...,{\bf{a}}_t^i,...,{\bf{a}}_t^{N}\} \in \mathbb{R} ^{N\times d}$ denotes the agent node representation at time $t$ and $d$ is the dimension of each node. ${\bf{E}}^{HRG}_t$ is the cross-agent risk edge matrix. $S_t$ is the road semantic layout at time $t$. Notably, the agent $i$ can come from different agent categories, such as pedestrians, cars, or cyclists.

\subsubsection{Node Representation (NR) of HRG}

\begin{figure}[!t]
  \centering
 \includegraphics[width=\hsize]{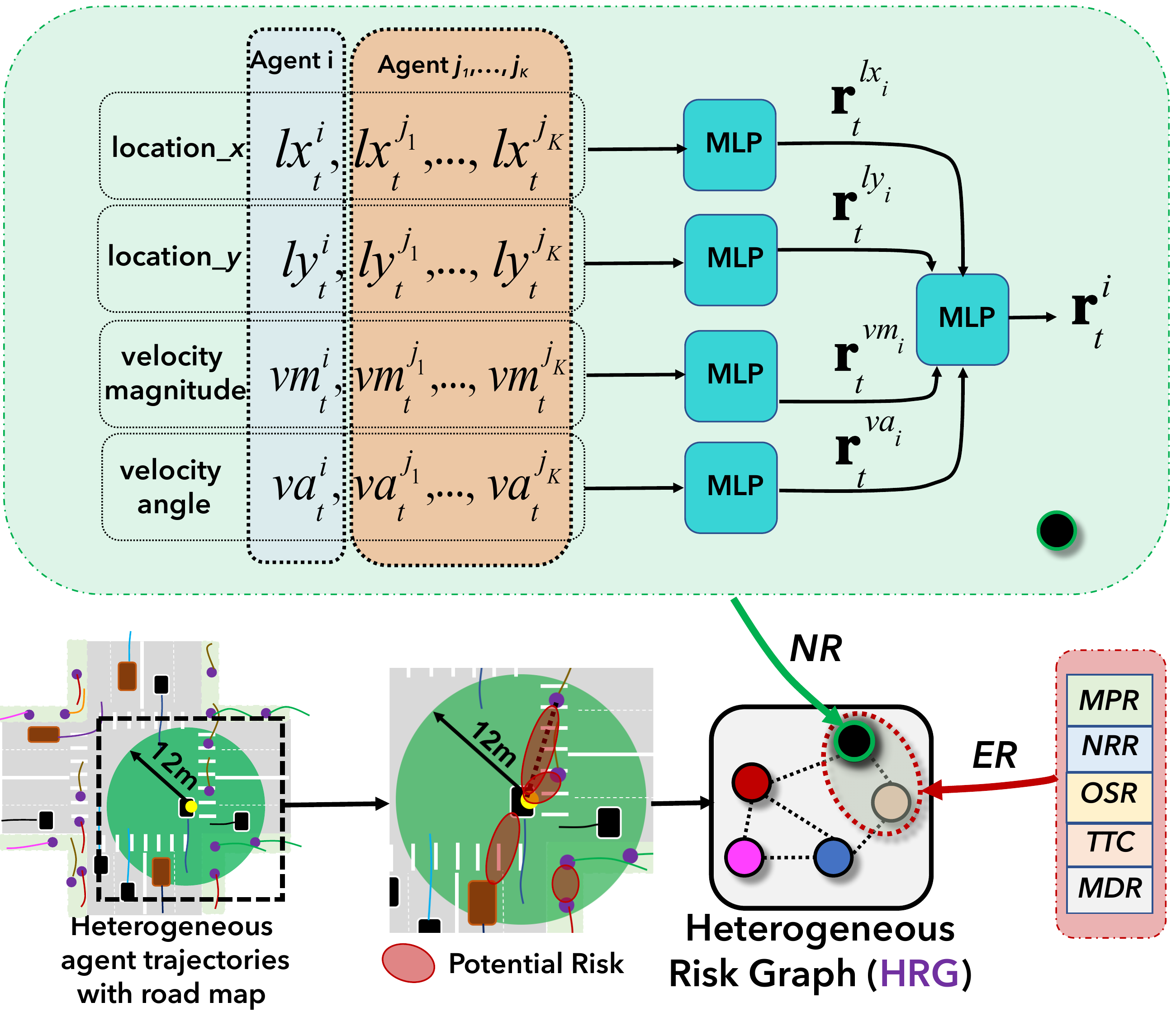}
  \caption{The construction of HRG. The abbreviations in this figure are  \textbf{MPR}: moving pattern relation; \textbf{NRR}: node representation relation; \textbf{OSR}: occupied semantic relation;\textbf{TTC}: time to collision; \textbf{MDR}: moving direction relation; \textbf{NR}: node representation and \textbf{ER}: edge representation.}
  \label{fig3}
\end{figure}

For a road agent, its representation is commonly modeled by the location, moving direction, and velocity. In addition, the influence of the agent's neighborhood takes an important role to restrict their moving trend. In this work, we embed this information in a pure and simple Multiple Layers of Perception (MLP) architecture, which is lightweight for further implementation. The node representation of HRG is illustrated in Fig. \ref{fig3}, and the representation process can be defined as
\begin{equation}
{\bf{r}}_t^{att_i}=\text{MLP}(att_t^{i},att_t^{j_{1}},...,att_t^{j_{k}},...,att_t^{j_{K}}),
\label{eq:2}
\end{equation}
where \text{MLP}() denotes MLP constructed by three groups of fully-connected layers and \emph{Relu()} function. ${\bf{r}}^{att_i}\in \mathbb{R}^{1\times 128}$ denotes the attribute embedding of the location ($lx, ly$), the velocity magnitude $vm$, the velocity angle $va$ of agent $i$, and the attribute relation between agent $i$ and its neighborhood agents. $att_t^{i}$ and $att_t^{j_{K}}$  are the attribute of agent $i$ and the $k^{th}$ agent $j$ at time $t$, respectively. $K$ specifies the number of neighborhood agents of agent $i$. In this work, we set the radius of the neighborhood as $\textbf{12}$ meters. Based on the MLP module in Eq. \ref{eq:2}, we can aggregate the attributes of agent $i$ and its neighborhood agents no matter what $K$ is. 

With the embedding of each agent attribute, the node representation ${\bf{a}}^i_t$ of the agent $i$ is defined as
\begin{equation}
{\bf{a}}^i_t=\text{MLP}({\bf{r}}_t^{lx_i},{\bf{r}}_t^{ly_i},{\bf{r}}_t^{vm_i},{\bf{r}}_t^{va_i}),
\label{eq:3}
\end{equation}
where ${\bf{a}}^i_t$ is a 64-dimensional vector. 
\subsubsection{Edge Representation (ER) of HRG}
For the collision risk between road agents, their distance, moving patterns (e.g., ``\emph{straight walking}", ``\emph{running}", ``\emph{crossing}", and ``\emph{stopping}" of pedestrians, ``\emph{turning}", ``\emph{accelerating}", and ``\emph{moving with constant speed}" of vehicles), and movable road regions (e.g., sidewalk for persons, and road region for vehicles) are all the key factors for modeling the collision risk. Therefore, we model the edge representation of HRG from the cross-relation between the node representations, moving patterns, occupancy of the road semantic types, moving direction, and the distance of different agents. Notably, we take the distance and moving direction of road agents again because the collision risk has the most direct link with them.

Consequently, the collision risk metric between agent $i$ and agent $j$ is defined as
\begin{equation}
\begin{array}{ll}
e^{ij}_t=\mathbb{I}(s^i_t=s^j_t)*O^{ij}_t*{\frac{1}{|T_{t}^{ij}|}}*m^{ij}_t,\\
s.t., s^i_t, s^j_t\in S_t,\\
\end{array}
\label{eq:4}
\end{equation}
where $e^{ij}_t \in {\bf{E}}_{t}^{HRG}$ is a scalar for risk evaluation between agent $i$ and agent $j$, and $|T_{t}^{ij}|$ represents the moving time when agent $i$ and $j$ may collide in the interaction. 

\textbf{Occupied Semantic Relation (OSR)}: $\mathbb{I}(s^i_t=s^j_t)$ represents an indicator function that is set as 0 if the occupied road semantic types of the agent $i$ and agent $j$ are different (e.g., sidewalk for pedestrians and road for vehicles), and set to 1 for the agents that stand in the same road semantic type (e.g., pedestrians and cars all move in the road region). Notably, for the zebra and road regions, their semantic labels are denoted as the same in HRG.

\begin{figure}[!t]
  \centering
 \includegraphics[width=0.65\hsize]{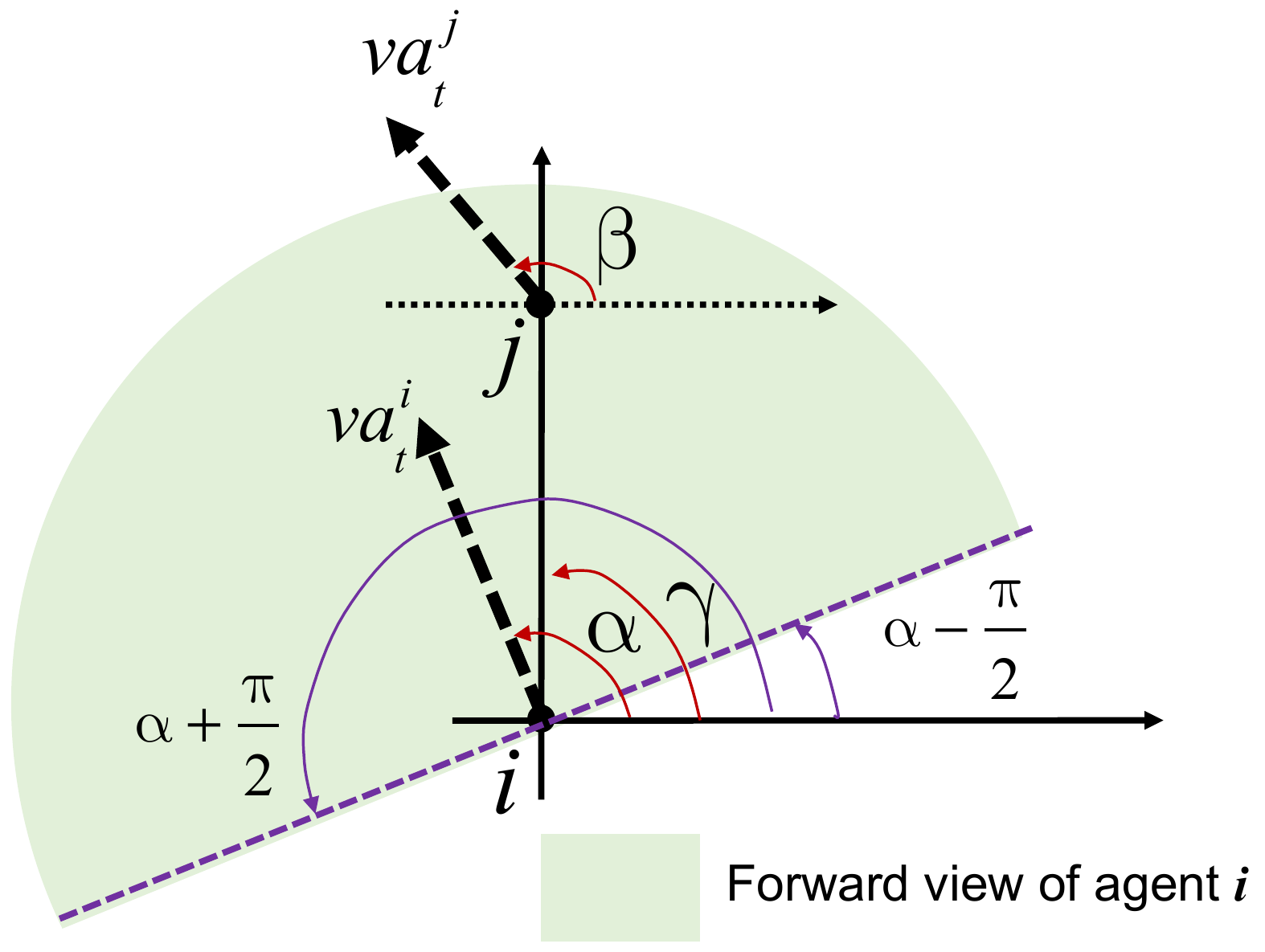}
  \caption{The moving direction relation between agent $i$ and $j$.}
  \label{fig4}
\end{figure}

\textbf{Time to Collision (TTC)}: Since the TTC is inversely proportional to the collision risk, ${1/|T_{t}^{ij}|}$ is used to represent the collision risk coefficient, and $T^{t}_{ij}$ is defined as
\begin{equation}
|T_{t}^{ij}|=\frac{||{\bf{p}}^i_t-{\bf{p}}^j_t||_2}{\left|{vm^i}_t*cos(\left|\alpha-\gamma\right|)-vm^j_t*cos(\left|\beta-\gamma\right|)\right|},\\
\end{equation}
where  ${\bf{p}}^i_t$ and ${\bf{p}}^j_t$ denote the locations of agent $i$ and agent $j$ at time $t$, respectively.  $\alpha$ and $\beta$ are the moving angles of the agent ${i}$ and ${j}$ with respect to the horizontal axis. $\gamma$ denotes the angle between the linking lane of agent ${i}$ and agent ${j}$ with the horizontal axis, as illustrated in Fig. \ref{fig4}.

After the above collision risk modeling, the main insight of this work is to involve the Moving Pattern Relation (\textbf{MPR}) and the Node Representation Relation (\textbf{NRR}) between road agents, \emph{i.e.}, modeling $m^{ij}_t$. From the node representation, we can obtain the node representation ${{\bf{a}}^i_t}$ and ${{\bf{a}}^j_t}$ by Eq. \ref{eq:3} for agent $i$ and agent $j$. For moving pattern $C_t^{i}$ and $C_t^{j}$ of agent $i$ and agent $j$, we group the target trajectory points of observation of different kinds of road agents by spectral clustering method \cite{DBLP:conf/nips/NgJW01} which is fulfilled by learning a graph with the node representation of linked trajectory points and their kernel distance ($e^{-\frac{||{\bf{p}}^i-{\bf{p}}^j||^2}{2\sigma})}$). The kernel distance of all the trajectory samples constructs the affinity matrix ${\bf{W}}$, and the graph is optimized by searching the minimum cut of the standard Laplacian matrix ${\bf{D}}^{-\frac{1}{2}}{\bf{LD}}^{-\frac{1}{2}}$, where ${\bf{D}}$ is the degree matrix of ${\bf{W}}$ and ${\bf{L}}={\bf{D}}-{\bf{W}}$. By the clustering process, we can obtain $C_t^{i}$ and $C_t^{j}$. Then, $m^{ij}_t$ is denoted as
\begin{equation}
m^{ij}_t=\text{MLP}({{\bf{a}}^i_t}, {{\bf{a}}^j_t}, C_t^{i},C_t^{j}),\\
\end{equation}
where $C_t^{i}$ and $C_t^{j}$ represent the cluster index of agent $i$ and agent $j$ corresponding to the moving patterns clustered by the trajectories, respectively. Notably, the structure of MLP is similar to the one defined in Eq. \ref{eq:3} but with an output dimension of $1$.

\textbf{Moving Direction Risk (MDR)}: If the surrounding agents of agent $i$ appear in its forward view (marked by green color in  Fig. \ref{fig4}), the risk coefficient is set as 1, and vice versa (0), i.e.,
\begin{equation}
O^{ij}_t=\left\{
\begin{aligned}
1       &      & \gamma \in [\alpha-\frac{\pi}{2}, \alpha+ \frac{\pi}{2}],\\
0       &   & {else},\\
\end{aligned}
\right.
\end{equation}
where the meaning of $\alpha, \beta$, and $\gamma$ is demonstrated in Fig. \ref{fig4}.

\subsection{Hierarchical Scene Graph (HSG)}
\label{rssg}

In mixed traffic scenes, the movements of various road agents are affected by the semantic layout and the road utility to some extent. For example, pedestrians commonly move on the sidewalk and vehicles move on the main road in the actual driving, where the movable road regions. Many previous works exploit the road semantic type and adopt the semantic image directly into the scene feature embedding \cite{DBLP:conf/eccv/SalzmannICP20}, which is not explainable and shows an unsatisfying performance gain for trajectory forecasting. 

The road scene has a natural grammar structure prior. For example, road region links with the drivable area, carpark, walkway, etc. This kind of grammar structure of road semantic types is called road scene graph \cite{DBLP:conf/icra/PrakashBBACSSB19,DBLP:journals/kbs/MalawadeYHKKF22}, and scene graph models are successfully adopted in video grounding \cite{DBLP:conf/eccv/XiaoZCY22}, synthetic data generation \cite{DBLP:conf/iros/SavkinENT21}, and so on. To leverage this prior knowledge of road structure, we model the road scene graph for trajectory forecasting. Although the scene graph is called ``graph", its structure is like a ``tree" structure. Motivated by the work for virtual scene simulation \cite{DBLP:conf/iccv/KarPLCYRA0F19,DBLP:conf/eccv/DevaranjanKF20}, the scene graph structure is shown in Fig. \ref{fig5}, which is a hierarchical structure expression. This structure is universal for road scenes, and we take it as a basis $\mathcal{G}_{basis}$ for the hierarchical scene graph construction for each frame in the trajectory forecasting.

To be specific, assume the hierarchical scene graph for the $t^{th}$ frame is denoted as $\mathcal{G}^{HSG}_t=({\bf{V}}_t, {\bf{E}}^{HSG}_t)$, where ${\bf{V}}_t$ represents the set of semantic items in the frame, and ${\bf{E}}^{HSG}_t$ specifies the linking relation in $\mathcal{G}^{HSG}_t$.  ${\bf{v}}_t^p \in {\bf{V}}_t$ is the $p^{th}$ semantic item in the $t^{th}$ frame, including road agents and static semantic regions.  In the following, we will describe the construction process of HSG.

\subsubsection{Modeling ${\bf{V}}_t$}
In the traffic scene, the semantic items of the scene graph are represented by the coordinate of each kind of semantic region. For the road agent, the center point (meters) of the 3D point cloud is taken (in this work, the height value is omitted). For the semantic regions of drivable regions, sidewalks, and other static semantic regions, the 2D center points of these regions surrounded by various polygons over the ground plane are adopted. Consequently, for the vertex of $\mathcal{G}^{HSG}_t$, the dimension of ${\bf{V}}_t$ is also denoted as $\mathbb{R}^{(N+M)\times 2}$, where $N$ is the numbers of the road agents and $M$ denotes the static road semantic regions. In this work, we consider three kinds of road agents, \emph{i.e.},

-- ``\emph{pedestrians}", ``\emph{cars}" and ``\emph{riders}",

and eight categories of static road semantic types, \emph{i.e.},

-- ``\emph{road segment}", ``\emph{drivable area}", ``\emph{zebra region}",``\emph{carpark}", ``\emph{sidewalk}", ``\emph{road block}", ``\emph{road lanes}", and ``\emph{stop lines}".

\begin{figure}[!t]
  \centering
 \includegraphics[width=\hsize]{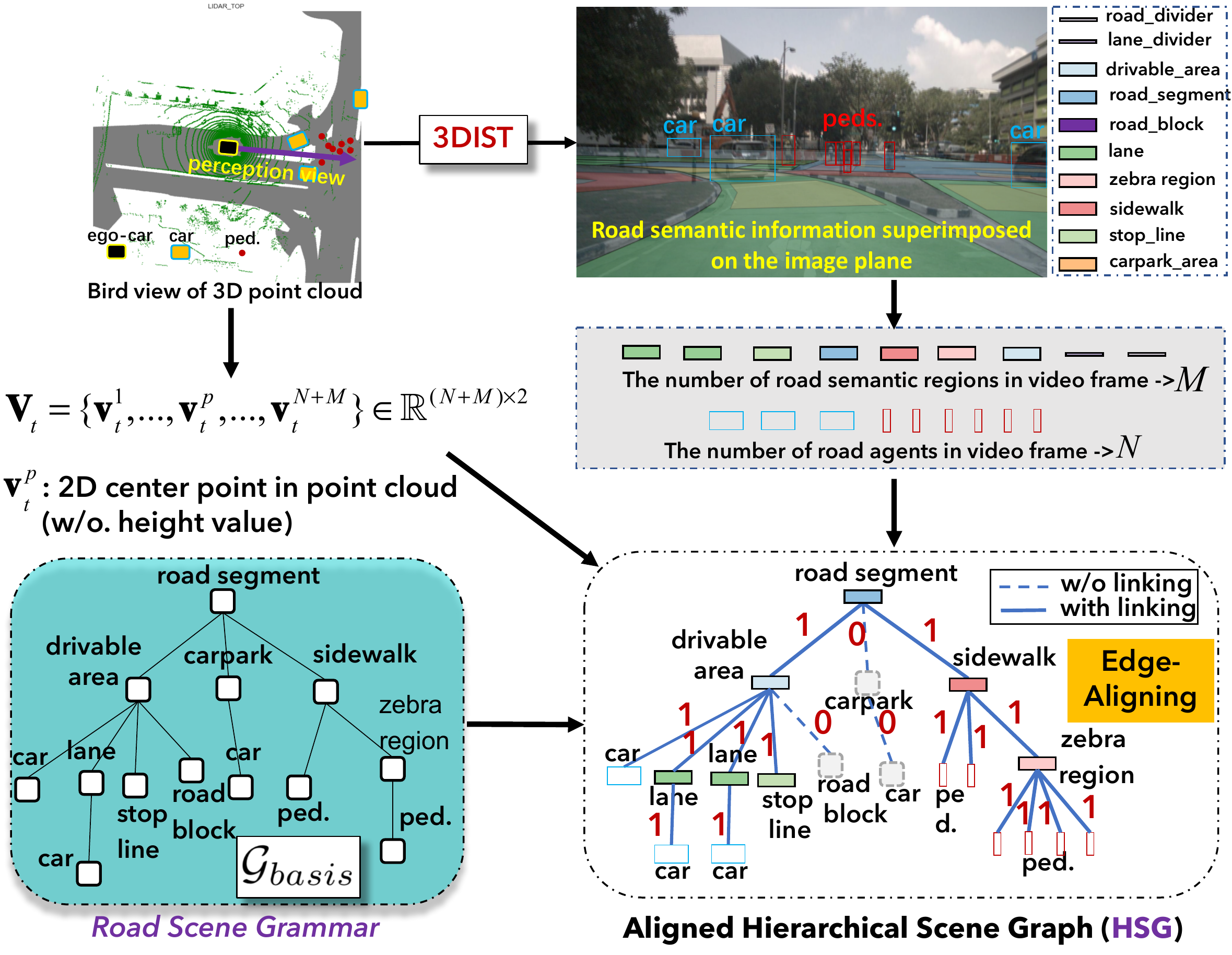}
  \caption{The construction process of HSG, where 3DIST denotes the transformation from the world coordinate system to the image coordinate system. Edge-Aligning is fulfilled by a one-hot operation on the linking of road agents and road semantic types supervised by the road scene graph. The solid lines in HSG represent the activated edge correlation, while the dashed lines specify the omitted edge correlation (best viewed in color mode).}
  \label{fig5}
\end{figure}

An example of this road semantic layout is presented in Fig. \ref{fig5}. Notably, there may be several regions with the same semantic type in each frame, which makes $M$ in each frame different. To be feasible for the following graph embedding, we set $M$ as max($M_1,...,M_{T_{obs}}$) over $T_{obs}$ frames.

\subsubsection{Modeling ${\bf{E}}_t^{HSG}$}
The edge matrix ${\bf{E}}_t^{HSG}$ in $\mathcal{G}^{HSG}_t$ correlates the natural link relation between different road semantic items. The modeling of the scene graph needs to pre-annotate the road semantic regions, and there should be available annotated road semantic regions in the Bird's Eye View (BEV) of the 3D point cloud. As aforementioned, we take the center points of road agents and the static semantic regions as the nodes of $\mathcal{G}^{HSG}_t$. If we model the scene graph edge in the 3D point cloud, there are vast linking relations in the large range of the 3D point cloud, which is time-consuming for constructing $\mathcal{G}^{HSG}_t$ and influences the use of scene graph. In addition, for HSG, we need to correlate the frame-level semantic layout with the trajectory points of agents, which needs a restricted range for semantic region consideration. The whole semantic information in the 3D point cloud is not appropriate manifestly. 

In this work, we model the edge linking of $\mathcal{G}^{HSG}_t$ by checking the relative relation between the road semantic region without needing an accurate practical physical distance. If the specific semantic type appears in the $t^{th}$ frame, we consider it and count the number of the regions with the same semantic type. Because of the large range of the 3D point cloud, we project the semantic layout of the 3D point cloud into an RGB image plane, which can restrict the calculation range to the forward view and reduce the number of the nodes in $\mathcal{G}^{HSG}_t$. The transformation from 3D point cloud to RGB semantic image plane is called 3DIST and is fulfilled by the transformation process of \emph{world coordinate system} $\rightarrow$\emph{camera coordinate system} $\rightarrow$ \emph{image coordinate system}. The transformation is done by the official code on the website of nuScenes dataset \footnote{https://github.com/nutonomy/nuscenes-devkit}. 

With this setting, we can use the road scene grammar (denoted as $\mathcal{G}_{basis}$)  to effectively construct $\mathcal{G}^{HSG}_t$ by checking the co-existence of road semantic types. Specifically, with $\mathcal{G}_{basis}$, the edge modeling of HSG is fulfilled by a \emph{Scene Graph Edge Aligning} module.

\textbf{Scene Graph Edge Aligning:} We assume ${\bf{E}}^{HSG}_t\in \mathbb{R}^{(N+M)\times (N+M)}$ is denoted as a matrix, where each element in ${\bf{E}}^{HSG}_t$ specifies the linking relation between $N+M$ nodes in $\mathcal{G}_{HSG}^t$. We model the element in ${\bf{E}}_t$  as a 0-1 encoding matrix, where the element with value 1 indicates the linking relationships between the relative semantic pair in the $t^{th}$ frame, and vice versa for the value of 0. We name this process as \textbf{Edge-Aligning} operation and fulfill by one-hot encoding for each pair of semantic items including road agents and static semantic regions.
\begin{equation}
{\bf{E}}_t^{HSG}=\textbf{Edge-Aligning}(\mathcal{G}_{basis},N,M, S'_t),
\end{equation}
where $S'_t$ is the semantic image after 3DIST from the original semantic annotation $S_t$ of 3D point cloud at time $t$. The detailed process of \textbf{Edge-Aligning} operation is conducted by checking the existence of the semantic pair of $\mathcal{G}_{basis}$ in $S'_t$, as illustrated in Fig. \ref{fig5}.

Thus, after the edge alignment of the road semantic regions at time $t$ on the basic scene graph $\mathcal{G}_{basis}$, we obtain the Hierarchical Scene Graph for the $t^{th}$ frame.

\subsection{Trajectory Forecasting}
\label{graphfusion}
With the Heterogeneous Risk Graph (HRG) $\mathcal{G}_{HRG}^t$ and the Hierarchical Scene Graph (HSG) $\mathcal{G}_{HSG}^t$ at time $t$, HRG and HSG are fed into a stacked Graph Convolution Network (GCN) and Temporal Convolution Network (TCN) \cite{DBLP:conf/cvpr/XuW0022} to encode the spatial-temporal structure evolution of semantic-induced trajectories of agents. Because of the same number of nodes in the time dimension, we stack the graph node representation of HRG and HSG up to time $t$ as ${\bf{A}}=\{{\bf{A}}_1,...,{\bf{A}}_{T_{obs}}\}$ and ${\bf{V}}=\{{\bf{V}}_1,...,{\bf{V}}_{T_{obs}}\}$, respectively. The stacked node graph embedding is denoted as ${\bf{H}}^{0}$. To provide the adaptive temporal dependency in the graph encoding, we take a residual connection in TCN. The formulation of graph encoding is
\begin{equation}
\begin{array}{rcl}
{\bf{H}}^{(l)}=\text{GCN}({\bf{H}}^{(l-1)}, {\bf{E}},{\bf{W}}_{gcn}^{(l-1)}),\\
{\bf{G}}=\text{TCN}({\bf{H}}^{(l)},{\bf{W}}_{t,tcn}^{(l)}),
\end{array}
\label{eq:8}
\end{equation}
where ${\bf{E}}$ is the stacked ${\bf{E}}_t^{HRG}$ or ${\bf{E}}_t^{HSG}$ over $T_{obs}$ frames. ${\bf{H}}^{(l)}$ is the output of the $l^{th}$ GCN layer ($l=3$). ${\bf{W}}_{gcn}^{(l-1)}$ is a learnable weight matrix of GCN layer, and ${\bf{W}}_{t,tcn}^{(l)}$ denotes the learnable parameters of TCN layer ($l=3$). ${\bf{G}}$ is the graph embedding for future trajectory prediction. TCN means temporal-CNN and models the dependency in the temporal dimension of ${\bf{H}}^{(l)}$ and predicts the future feature embedding in each layer ($l$), which is jointly optimized with GCN. For sufficiency, a concise description of GCN and TCN modules in Eq. \ref{eq:8} is provided.

\textbf{GCN:} GCN models the spatial representation correlation of different nodes \cite{DBLP:conf/cvpr/MohamedQEC20}. Different from traditional graph convolution operation, the interaction matrix ${\bf{E}}$ obtained by HRG or HSG can be treated as the affinity matrix. Therefore, the graph convolution operation in this work can be defined as ${\bf{H}}^{(l)}=\delta({\bf{D}}^{-\frac{1}{2}}_g({\bf{E}}+{\bf{I}}){\bf{D}}_g^{-\frac{1}{2}}{\bf{H}}^{(l-1)} {\bf{W}}_{gcn}^{(l-1)})$, where ${\bf{I}}$ is the identity matrix, ${\bf{D}}_g$ is the degree matrix computed from {\bf{E}}, and $\delta$(.) is the activated function, \emph{i.e.}, Parametric Rectified Linear Unit (PReLU) \cite{DBLP:conf/iccv/HeZRS15}.

\textbf{TCN:} TCN consists of three Residual Blocks (ResBs) \cite{DBLP:conf/cvpr/XuW0022}, and each ResB encodes the embedding after GCN and contains a temporal 1D-convolution, a batch normalization, a PReLU function, a dropout, and a residual link by 1D-convolution between the input and output of each ResB. TCN is fulfilled by $\eta({\bf{H}}^{(l)}{\bf{W}}_{t,tcn}^{(l)}+res)$, where $res$ denotes the residual link in TCN layer, and $\eta(.)$ is the activated function PReLU. 

\textbf{Graph Embedding Fusion:} To check the roles of HRG and HSG purely, we fuse the embedding of HSG and HRG with a simple dot-operation by 
\begin{equation}
\hat{{\bf{G}}}={\bf{G}}^{HRG}\odot \text{Conv}_{2D}({\bf{G}}^{HSG}),
\label{eq:9}
\end{equation}
where ${\bf{G}}^{HRG}\in\mathbb{R}^{N\times c\times T_{pred}}$, ${\bf{G}}^{HSG}\in\mathbb{R}^{(N+M)\times c\times T_{pred}}$, $\odot$ is the element-wise Hadamard product operation, and $c$ is the feature dimension and set as 5 in this work. 
We use a 2D-convolution operation to reduce the node dimension from $N+M$ to $N$, \emph{i.e.}, converting ${\bf{G}}^{HSG}$ to be with the same shape with ${\bf{G}}^{HRG}$.
We also compare another simple residual fusion by ${\bf{G}}^{HRG}\odot({\bf{I}}+\text{Conv$_{2D}$}({\bf{G}}^{HSG}))$ in the experiment (with a less role of HSG). After obtaining the fused graph embedding $\hat{{\bf{G}}}$, it is used to determine and decode the parameters in the following loss function. The decoding process is conducted by two temporal convolution layers, a PReLU activation function, and one linear layer, where the convolution kernel size is $3\times3$.

\textbf{Loss function:} In this work, we introduce the Bivariate Gaussian Probability Distribution (BGPD) \cite{DBLP:conf/cvpr/MohamedQEC20} to regress the predicted trajectories of heterogeneous agents, \emph{i.e.}, modeling $\mathcal{P}(.)$ in Eq. \ref{eq:1}. Hence, our model is trained by minimizing the following negative log-likelihood loss as
\begin{equation}
\begin{array}{ll}
\mathcal{L}=-\sum^{T_{obs}+T_{pred}}_{t=T_{obs}+1} \text{log} (\mathcal{P}({\bf{p}}_t^i|\hat{\mu}^i_t,\hat{\sigma}^i_t,\rho^i_t)), \\
s.t., \hat{{\bf{p}}}_t^i \sim \mathcal{N}(\hat{\mu}^i_t,\hat{\sigma}^i_t,\hat{\rho}^i_t),
\end{array}
\end{equation}
where $\hat{{\bf{p}}}_t^i$ is the predicted point of ${\bf{p}}_t^i$ at time $t$. $\hat{\mu}^i_t,\hat{\sigma}^i_t$, and $\hat{\rho}^i_t$ are the estimated mean, standard deviation, and correlation coefficient of $\hat{{\bf{p}}}_t^i$ by BGPD. $\mathcal{P}({\bf{p}}_t^i|\hat{\mu}^i_t,\hat{\sigma}^i_t,\hat{\rho}^i_t)$ is computed by
\begin{equation}
\begin{array}{ll}
\mathcal{P}({\bf{p}}_t^i|\hat{\mu}^i_t,\hat{\sigma}^i_t,\hat{\rho}^i_t)=\frac{1}{2\pi \hat{\sigma}^{i,x}_t \hat{\sigma}^{i,y}_t \sqrt{1-({\hat{\rho}^{i,xy}_t}})^2}\\
\cdot \exp(-\frac{1}{2(1-\hat{\rho}^{i,xy}_t)}[(\frac{\hat{x}_t^i-\hat{\mu}^{i,x}_t}{\hat{\sigma}^{i,\hat{x}}_t})^2+(\frac{\hat{y}_t^i-\hat{\mu}^{i,y}_t}{\hat{\sigma}^{i,y}_t})^2\\
-2\hat{\rho}^{i,xy}_t\frac{(\hat{x}_t^{i}-\hat{\mu}^{i,x}_t)(\hat{y}_t^{i}-\hat{\mu}^{i,y}_t)}{\hat{\sigma}^{i,x}_t\hat{\sigma}^{i,y}_t}]),
\end{array}
\end{equation}
where ($\hat{\mu}^{i,x}_t,\hat{\mu}^{i,y}_t, \hat{\sigma}^{i,x}_t, \hat{\sigma}^{i,y}_t$, and $\hat{\rho}^{i,xy}_t$) is estimated by the predicted point $\hat{{\bf{p}}}_t^i=(\hat{x}_t^i,\hat{y}_t^i)$ at time $t$ in the training, and $(\hat{\mu}^{i,x}_t,\hat{\mu}^{i,y}_t)$ is denoted by the ground-truth ${\bf{p}}_t^i=(x_t^i,y_t^i)$ in loss computation. Consequently, the value $c$ in Fig. \ref{fig2} is 5.

\section{Experiments and Discussions}
\label{expe}
\subsection{Datasets}
In this work, we take three challenging datasets to evaluate the proposed method. They are nuScenes \cite{DBLP:conf/cvpr/CaesarBLVLXKPBB20}, Argoverse \cite{DBLP:conf/cvpr/ChangLSSBHW0LRH19}, and ApolloScape \cite{DBLP:journals/pami/HuangWCZGY20} datasets. The detailed information of each dataset is as follows.

\textbf{nuScenes} is constructed by nuTonomy and released in 2019. It consists of many perception tasks in autonomous driving, including 3D object tracking, detection, trajectory forecasting, and 3D point semantic segmentation. nuScenes has 1000 driving sequences collected in Boston and Singapore, and each sequence takes 20 seconds and the annotation frame rate is 2Hz. In other words, there are two annotated frames in one second. nuScenes labels the 3D bounding boxes for 23 kinds of objects and has 11 categories of static road semantic regions. In this work,  the data for the trajectory forecasting task in nuScenes is taken.

\textbf{Argoverse} is developed by Argo AI, Carnegie Mellon University, and Georgia Institute of Technology, which focuses on 3D car tracking and motion forecasting tasks. There are LiDAR point clouds, RGB videos, and a high-accurate road map (HD map) with 290km. The motion forecasting part in Argoverse collected 763 hours of city scenarios, which were divided into 250,000 clips of 11 seconds, and the data frame rate is 10Hz. Notably, there are only vehicles in this dataset.

\textbf{ApolloScape} is released by Baidu in 2018. Similarly, it also contains the tasks of 3D detection, tracking, trajectory prediction, and semantic segmentation. There are 100k labeled images and 80k frames of 3D point cloud collected from 1000 km in the urban city. Differently, ApolloScape consists of 53 sequences and each one lasts 1 minute. The annotation frame rate is the same as one of the nuScenes (2Hz). There are three kinds of road agents, \emph{i.e.}, pedestrians, vehicles, and riders. However, there is no high-accurate map in the ApolloScape dataset. Therefore, we evaluate the performance of the Heterogeneous Risk Graph on this dataset.

\begin{table}[!t]\footnotesize
  \centering
  \caption{Dataset characteristics. NC.: number of clips; NPT.: number of pedestrian trajectories; NVT.:number of vehicle trajectories; NRT.: number of rider trajectories; all-T.: all trajectories.}
  \setlength{\tabcolsep}{1.4mm}{
\begin{tabular}{c|c|c|c|c|c|c}
\toprule[0.8pt]
datasets &NC.&NPT.&NVT.&NRT.&all-T. & training/testing\\
 \hline
nuScenes& 1000& 18,085& 53,217 &  1,296 & 63,785& 38,271/1,2756\\
ApolloScape & 53& 15,944& 6,274  & 4,495 & 26,713 &  16,028/5,343\\
Argoverse & 113& -& 324k  & - & 324k &  194k/64k\\
\toprule[0.8pt]
  \end{tabular}}
  \label{tab1}
  \end{table}
The detailed information of three datasets for training and testing setting is shown in Table. \ref{tab1}. This table shows that different datasets have diverse ratios of trajectory distribution for various road agents. It is worth noting that the riders take the smallest part in the datasets.

 \begin{figure}[!t]
  \centering
 \includegraphics[width=\hsize]{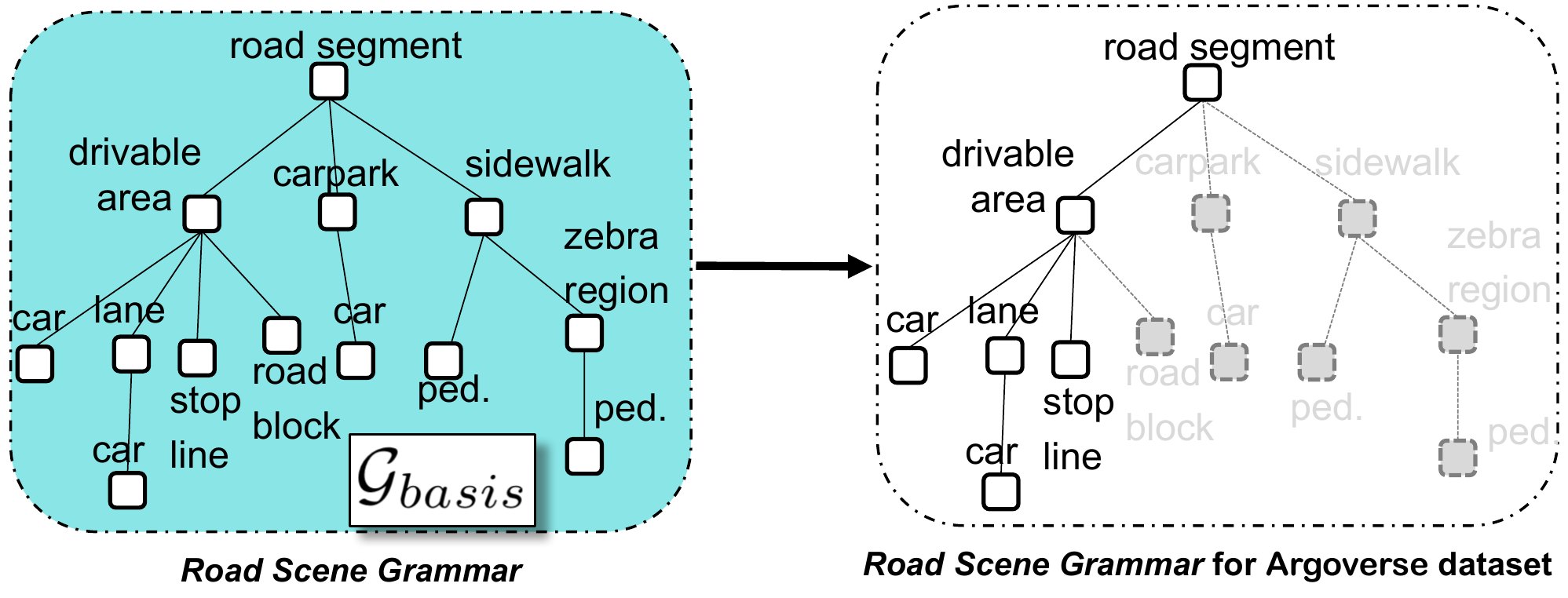}
  \caption{The road scene grammar graph in the Argoverse dataset.}
  \label{fig6}
  \vspace{-1em}
\end{figure}

\subsection{Implementation Details}

Following other works, most of the experiments are with 2 seconds of observation and 3 seconds (@3s) of prediction in these datasets. For nuScenes, we also provide a 6s-prediction (@6s) case in this work. For each dataset, the training epoch is set as 50, and the learning rate is initialized as 0.001 with a decreasing ratio of 0.2 for every 5 epochs. The kernel size of the convolution in GCN is 3, and the batch size of trajectories is 1024. Adam optimizer is used in loss function optimization.

Notably, we adopt the spectral clustering method to obtain the moving patterns of the trajectories. Because different kinds of road agents own differing maneuverability, we set the clusters of pedestrians, vehicles, and riders are 6, 3, and 3, respectively. All experiments are conducted on a platform with a 2080Ti GPU and 64G RAM. The code of this work is released on the website of \footnote{\url{https://github.com/JWFanggit/HRG_HSG_trajForecasting}}.

Because the HD map of the Argoverse dataset only provides the road structure, it does not have the fine-grained semantic layout of the road scene, such as the sidewalk, roadblock, etc. Therefore, the road scene grammar of HSG for the Argoverse dataset is simplified as Fig. \ref{fig6}. However, to maintain the same data input type, we also take the whole structure of road scene grammar and treat HSG's edge value of the unrelated road semantic relation as zero. In addition, with the limited semantic annotation, we cannot project the HD map to the RGB images like the ones of the nuScenes dataset. Therefore, for the Argoverse dataset, we obtain the semantic relation between the cars and road region by checking the semantic regions occupied by the car in the 3D point clouds. In other words, the HSG of the Argoverse dataset is car-centric. 
 \begin{figure*}[!t]
  \centering
 \includegraphics[width=\hsize]{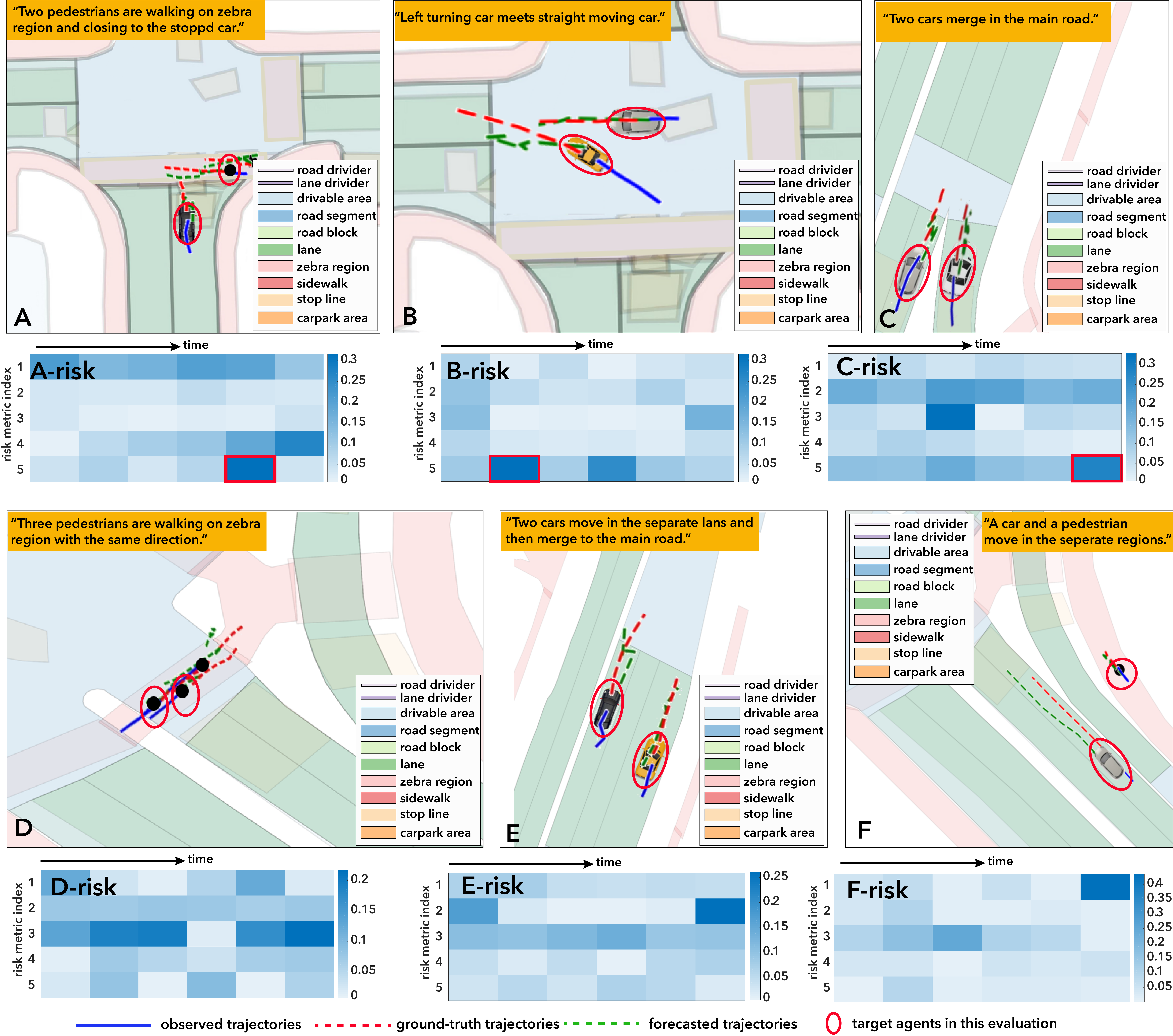}
  \caption{Some typical situations of forecasted trajectories in nuScenes dataset with the evaluation of different combinations of risk metrics (The pair of road scenes and the risk matrix are shown). Among the risk metric index, from the $1^{st}$to the $5^{th}$ risk metric index noted in the vertical axis of the risk matrix, the combination of the risk metrics are NRR, NRR+MPR, NRR+TTC, NRR+MPR+TTC+MDR, and NRR+MPR+TTC+MDR+OSR. The horizontal axis represents the $6$ future frame index (@3s prediction).}
  \label{fig7}
\end{figure*}

\subsection{Metrics}
The metrics in this work take the popular Average Distance Error (ADE) and Final Distance Error (FDE), defined as
\begin{equation}
\begin{array}{rcl}
\text{ADE}=\frac{1}{N}\sum_{i=1}^N \sum_{t=T_{obs+1}}^{T_{obs}+T_{pred}} ||\hat{{\bf{p}}}_t^i-{\bf{p}}_t^i||_2,\\
\text{FDE}=\frac{1}{N}\sum_{i=1}^N ||\hat{{\bf{p}}}_{T_{obs}+T_{pred}}^i-{\bf{p}}_{T_{obs}+T_{pred}}^i||_2,
\end{array}
\end{equation}
where $\hat{{\bf{p}}}_{t}^i$ and $\hat{{\bf{p}}}_{T_{obs}+T_{pred}}^i$ are the predicted trajectory points and the last one, respectively. To avoid the one-time occasionality in trajectory forecasting, we repeat $h=$20 times of prediction for each version of the model, and the predictions with the minimum ADE (\textbf{mADE}) and minimum FDE (\textbf{mFDE}) are selected as the final results. In addition, Missing Rate (\textbf{MR}$_{2,h}$) computes the ratio of predicted points in the testing set whose mFDE is greater than 2 meters with $h$ times of prediction. The smaller MR means better performance.

Notably, to avoid the influence of one round of tests, the final ADE and FDE are obtained by selecting the minimum in five test trials. Especially, for the ApolloScape dataset, it provides an official weighted ADE and FDE restricting the unbalancing of the pedestrians ($p$), vehicles ($v$), and riders ($r$) in evaluation, denoted as
\vspace{-0.1em}
\begin{equation}\small
\begin{array}{rcl}
\text{wADE}=0.2* \text{mADE}_{v}+0.58*  \text{mADE}_{p}+0.22*  \text{mADE}_{r}\\
\text{wFDE}=0.2*  \text{mFDE}_{v}+0.58*  \text{mFDE}_{P}+0.22* \text{mFDE}_{r}.
\end{array}
\end{equation}
\vspace{-2.5em}

\subsection{Ablation Studies}
\subsubsection{Influence of Different Risk Metrics}
\label{rmd}

In this work, we exploit five kinds of metrics for reasoning the risk in Eq. \ref{eq:4}, \emph{i.e.}, the Occupied Semantic Relation (OSR), Time to Collision (TTC), Moving Pattern Relation (MPR),  Moving Direction Risk (MDR), and the Note Representation Relation (NRR). Among them, because NRR computes the node representation distance in the edge modeling of HRG, it is taken as the baseline for checking the role of these metrics. For other metrics, we gradually fuse them from moving time, moving direction, road semantic constraint, and moving pattern, and evaluate the performance on the nuScenes dataset. The mADE and mFDE are adopted for evaluation. The evaluation results are shown in Table. \ref{tab2}.  

From the Table. \ref{tab2}, with more consideration of these risk metrics, the performance is boosted manifestly. Among them, NRR, MPR, and TTC reveal an equivalent level of forecasting. It is because, in the node representation of Heterogeneous Risk Graph (HRG), the velocity and location of agents are also considered. TTC and MPR are computed by the same information of the target agent. Therefore, the MPR and TTC have little impact on NRR in this evaluation. In addition, we find that introducing MDR and OSR to NRR has a significant influence on the final prediction performance. MDR and OSR have the apparent boundary of sensitivity, e.g., that they are activated when a similar moving direction or the same semantic region is taken by different agents. 

\begin{table}[!t]\small
  \centering
  \caption{Influence of different risk metrics in HRG on the testing set of nuScenes dataset.}
\begin{tabular}{ccccc|c|c}
\toprule[0.8pt]
NRR&MPR&TTC&MDR&OSR& mADE $\downarrow$ &mFDE $\downarrow$ \\
 \hline
 \checkmark& & &   &  &   1.21&1.96  \\
 \checkmark  & \checkmark & &   &  &   1.20& 1.93 \\
 \checkmark& \checkmark &  \checkmark&   &  &   1.19&  1.91\\
  \checkmark& \checkmark &  \checkmark&  \checkmark &  &   1.13&  1.66\\
    \checkmark& \checkmark &  \checkmark&  \checkmark &\checkmark  &   \textbf{1.09}&\textbf{1.56}  \\
\toprule[0.8pt]
  \end{tabular}
  \label{tab2}
  \end{table}
Besides, we also present some frameshots for visualizing the role of these metrics in Fig. \ref{fig7}.  From the results, considering all risk metrics (\emph{i.e.}, NRR+MPR+TTC+MDR+OSR) shows more reasonable risk evaluation results for the future trajectories. It is worth noting that from the risk matrix Fig. \ref{fig7} (C-risk), the situation of ``\emph{two cars merge in the main road}" obtains the increased risk with time, and the largest risk appears in the last frame, marked by red color, which is reasonable from the visualized road scene of Fig. \ref{fig7} (C). Similarly, from Fig. \ref{fig7} (B), and Fig. \ref{fig7} (B-risk), these two cars produce a large risk in the second future frame because they meet immediately and merge to the same main road. Therefore, HRG provides useful help for heterogeneous trajectory prediction.

\subsubsection{Contribution of HRG and HSG}

The main contribution of this work is to model Heterogeneous Risk Graph (HRG) and Hierarchical Scene Graph (HSG) to infer the heterogeneous trajectory forecasting problem. Their role are evaluated in this subsection. In addition, since the embedding of HRG and HSG are combined by a dot fusion strategy, defined in Eq. \ref{eq:9}, we also check the fusion strategy by comparing the residual fusion (\textbf{HRG+HSG-res.}) with the fusion strategy of dot-product operation (\textbf{HRG+HSG-dot.}). The performance is obtained by testing each combination on the nuScenes and Argoverse datasets, and the results are shown in Table. \ref{tab3}. It can be observed that HRG plays a significant role in trajectory forecasting.
The HRG+HSG-res. version and HRG demonstrate similar performance, while HRG+HSG-dot. generates the best results than others. It indicates that the scene graph learning in this work shows manifest promotion. 
 
  \begin{table}[htpb]\small
  \centering
  \caption{Performance comparison of HRG and HSG (meters).}
\begin{tabular}{c|c|c|c|c}
\toprule[0.8pt]
 \multirow{2}[4]{*}{Baselines} & \multicolumn{2}{c|}{nuScenes} & \multicolumn{2}{c}{Argoverse}\\
\cmidrule{2-5}          & mADE$\downarrow$ &mFDE$\downarrow$ & mADE$\downarrow$ & mFDE $\downarrow$ \\
 \hline
HRG  & 1.09& 1.560& 1.06 & 1.576\\
HRG+HSG-res. & 1.05 & 1.500& 1.05 & 1.356\\
HRG+HSG-dot.  & 0.93 & 1.320& 0.85 & 1.123  \\
\toprule[0.8pt]
  \end{tabular}
  \label{tab3}
  \end{table}
     \begin{figure}[!t]
  \centering
 \includegraphics[width=\hsize]{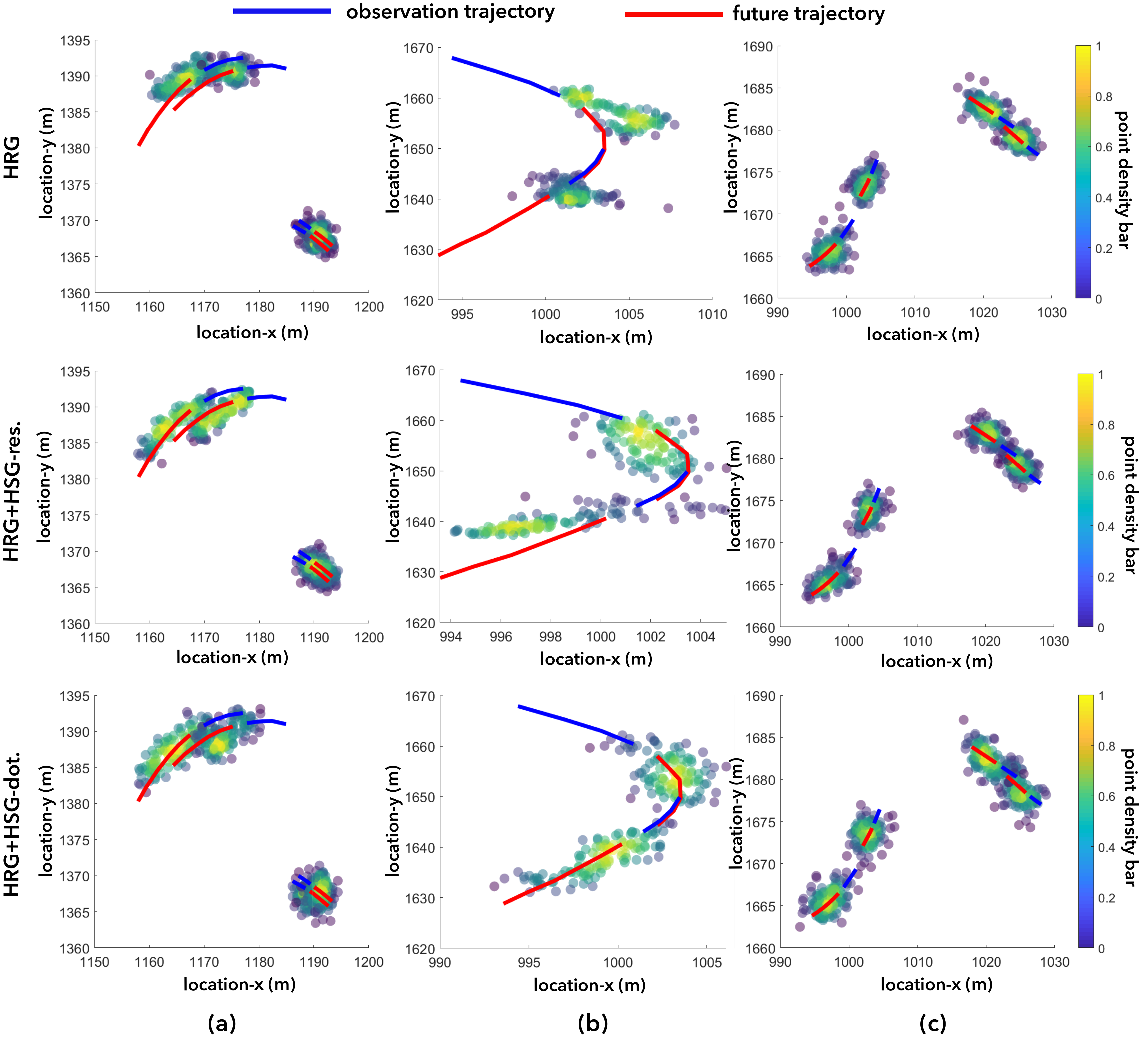}
  \caption{The trajectory point distribution of 20 times of @3s prediction for three samples in the nuScenes dataset. Notably, the behaviors in this figure are: (a) two cars are turning left and two persons are walking abreast, (b) two cars are turning right; (c) four persons are walking in a different direction.}
  \label{fig8}
\end{figure}

 \begin{figure}[!t]
  \centering
 \includegraphics[width=\hsize]{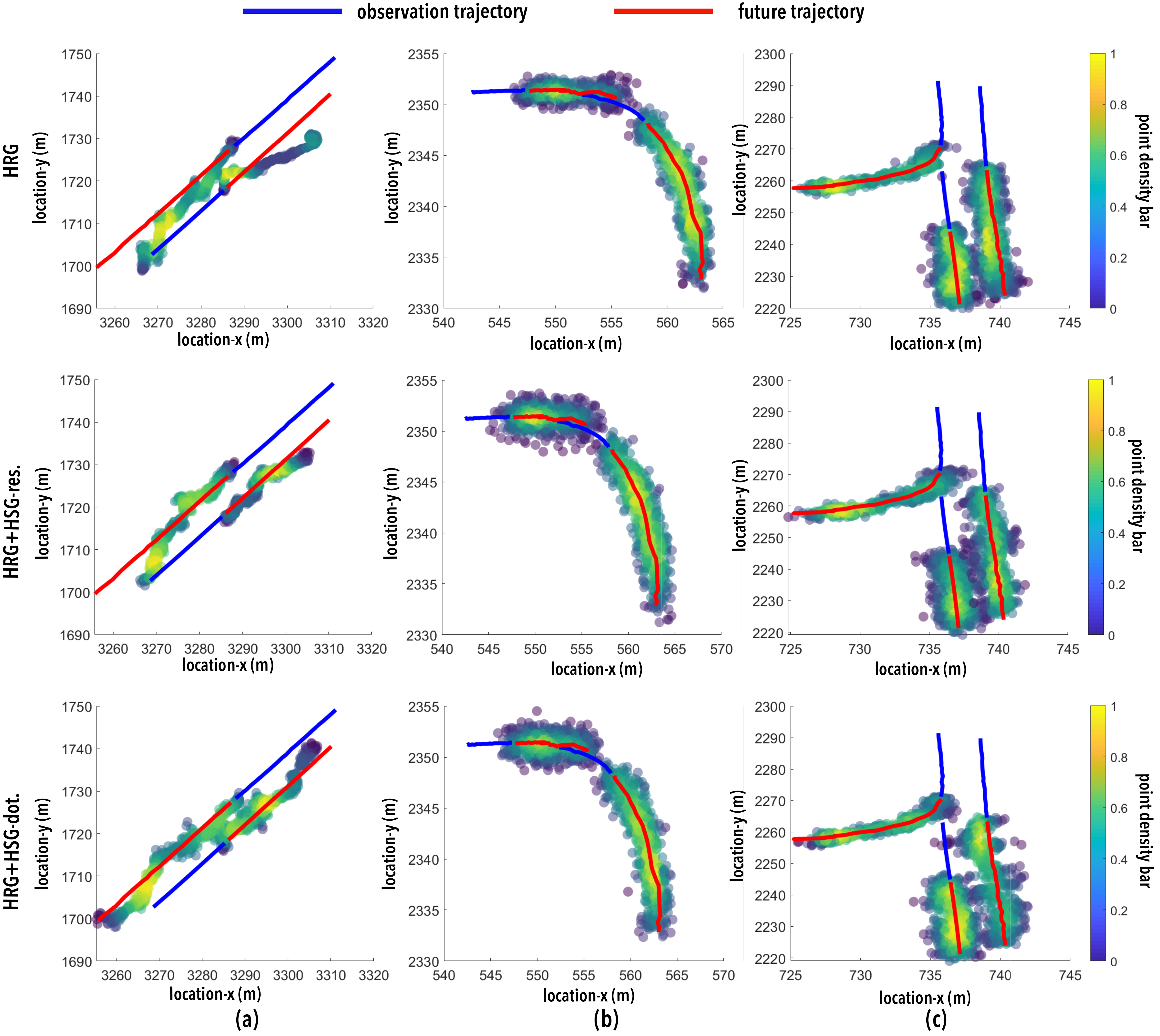}
  \caption{The trajectory point distribution of 20 times of @3s prediction for three samples in the Argoverse dataset. Notably, the behaviors in this figure are: (a) two cars are going straightly in opposite directions, (b) two cars are turning right; (c) two cars drive straightly and another one is turning right.}
  \label{fig9}
\end{figure}

\begin{table*}[htpb]\scriptsize
  \centering
  \caption{Performance comparison (@3s prediction) of the proposed method with other state-of-the-art approaches on the testing set of nuScenes and Argoverse datasets, where the extra configuration means extra information of scene except the raw trajectories.}
\begin{tabular}{c|c|c|c|c|c|c|c|c}
\toprule[0.8pt]
 \multirow{2}[4]{*}{Methods} & {extra information} & {encoder} & \multicolumn{2}{c|}{nuScenes}& \multicolumn{2}{c|}{Argoverse} & \multicolumn{2}{c}{Average}\\
 \cmidrule{4-9}   &     &   & mADE$\downarrow$ &mFDE$\downarrow$& mADE$\downarrow$ & mFDE $\downarrow$ & mADE$\downarrow$ & mFDE $\downarrow$\\
 \hline
Social-LSTM   \cite{DBLP:conf/cvpr/AlahiGRRLS16} $_{\scriptsize{\emph{CVPR2016}}}$&-&\textcolor{cyan}{LSTM}& 1.390 & 2.676&  1.385& 2.567&1.388&2.622\\
R2P2-MA \cite{DBLP:conf/eccv/RhinehartKV18}$_{\scriptsize{\emph{ECCV2018}}}$& -&CNN+\textcolor{cyan}{GRU}& 1.179& 2.194&1.108 & 1.771&1.143&1.983  \\
S-STGCNN \cite{DBLP:conf/cvpr/MohamedQEC20} $_{\scriptsize{\emph{CVPR2020}}}$& -&(STC-\textcolor{blue}{GCN})& 1.274 & 2.198&1.305 & 2.344&1.290&2.271\\
\hline
DESIRE* \cite{DBLP:conf/cvpr/LeeCVCTC17} $_{\scriptsize{\emph{CVPR2017}}}$&scene context image&\textcolor{cyan}{RNN}&1.079& 1.844 & 0.890 & 1.453 &0.985&1.649 \\
MFP \cite{DBLP:conf/nips/TangS19} $_{\scriptsize{\emph{NeuIPS2020}}}$ &scene context image&\textcolor{blue}{graphical} model+\textcolor{cyan}{RNN}& 1.301 &2.740& 1.399 & 2.684&1.350&2.712\\
MATFD \cite{DBLP:conf/cvpr/ZhaoXMCBZ0W19} $_{\scriptsize{\emph{CVPR2019}}}$& scene context image&conditional GAN+\textcolor{cyan}{LSTM}& 1.261 & 2.538& 1.344 &2.484&1.303&2.511\\
MATFG* \cite{DBLP:conf/cvpr/ZhaoXMCBZ0W19} $_{\scriptsize{\emph{CVPR2019}}}$& scene context image&conditional GAN+\textcolor{cyan}{LSTM}& 1.053 &2.126&  1.261 &2.313&1.157&2.219\\
Trajectron ++ \cite{DBLP:conf/eccv/SalzmannICP20} $_{\scriptsize{\emph{ECCV2020}}}$&road map image& \textcolor{magenta}{attention}+\textcolor{cyan}{GRU}&1.510 &n/a& n/a & n/a&n/a&n/a\\
TPNet \cite{DBLP:conf/cvpr/FangJSZ20} $_{\scriptsize{\emph{CVPR2019}}}$& road centerline& CNN & n/a&n/a& 1.610& 3.280&n/a&n/a\\
DATF \cite{DBLP:conf/eccv/ParkLSBKFJLM20} $_{\scriptsize{\emph{ECCV2020}}}$&scene context image&multimodal \textcolor{magenta}{attention}+\textcolor{cyan}{RNN}& 1.124 & 2.318 & 1.131 & 2.504 &1.127&2.411 \\
VectorNet \cite{DBLP:conf/cvpr/GaoSZSALS20} $_{\scriptsize{\emph{CVPR2020}}}$&road map&\textcolor{magenta}{attention}+MLP&1.520 & 2.890 & 1.660& n/a &1.590&n/a \\
CoverNet \cite{phan2020covernet} $_{\scriptsize{\emph{CVPR2020}}}$& road centerline & \textcolor{cyan}{LSTM}+CNN&1.480& n/a&  n/a & n/a&n/a&n/a\\
MultiPath \cite{DBLP:conf/corl/ChaiSBA19} $_{\scriptsize{\emph{CoRL2020}}}$& polygon road context & CNN+\textcolor{cyan}{RNN}&1.960& n/a& n/a & n/a&n/a&n/a\\
NLNI \cite{DBLP:conf/iccv/Zheng0ZTN0021}  $_{\scriptsize{\emph{ICCV2021}}}$& agent category & graph \textcolor{magenta}{attention}+(STC-\textcolor{blue}{GCN})& 1.049& 1.521& \textbf{0.792} & 1.256&0.921&1.388\\
ALAN \cite{DBLP:conf/cvpr/NarayananMPLC21} $_{\scriptsize{\emph{CVPR2021}}}$& road centerline & \textcolor{cyan}{LSTM}+CNN&n/a& 1.660& n/a & n/a&n/a&n/a\\
AgentFormer  \cite{DBLP:conf/iccv/0007WOK21} $_{\scriptsize{\emph{ICCV2021}}}$& road centerline & \textcolor{magenta}{Transformer}+MLP&1.450& 2.860&  n/a & n/a&n/a&n/a\\
PIP  \cite{DBLP:journals/corr/abs-2212-02181} $_{\scriptsize{\emph{arxiv2022}}}$& BEV+road map & \textcolor{magenta}{Transformer}&1.230& 1.750&  n/a & n/a&n/a&n/a\\
\hline
HRG+HSG-res.(ours)&risk+road semantic graph& (STC-\textcolor{blue}{GCN})& 1.050& 1.500& 1.050&1.356&1.050&1.428\\
HRG+HSG-dot.(ours)&risk+road semantic graph& (STC-\textcolor{blue}{GCN})& \textbf{0.930}& \textbf{1.320}& 0.850 &\textbf{1.123}&\textbf{0.890}&\textbf{1.221}\\
\toprule[0.8pt]
  \end{tabular}
  \label{tab4}
  \end{table*}
Because the results in the evaluation are obtained by selecting the forecasted results with the minimum ADE or FDE, we also visualized the 20 times of prediction results (future 3 seconds) of some typical samples of nuScenes in Fig. \ref{fig8}  and Argoverse datasets in Fig. \ref{fig9}  by HRG, HRG+HSG-res., and HRG+HSG-dot., respectively. From the predicted trajectory points, we can see that  ``HRG+HSG-res." boosts the long-term prediction compared with ``HRG", as shown in Fig. \ref{fig8} (a)  and Fig. \ref{fig8} (b). ``HRG+HSG-dot." has the best forecasting distribution than others, illustrated by Fig. \ref{fig8} (a), Fig. \ref{fig8} (b), and Fig. \ref{fig9} (a). As for the persons with small moving displacements, all versions of this work show similar performance in nuScenes. The density distribution of the predicted points and the results in Table. \ref{tab3} indicate that HSG provides a meaningful road scene context constraint for trajectory forecasting, which is promising in this domain.

\subsection{Comparison with the State-of-the-Art }
To verify the superiority of the proposed method, we compare it with many state-of-the-art approaches. They are Social-LSTM   \cite{DBLP:conf/cvpr/AlahiGRRLS16}, R2P2-MA \cite{DBLP:conf/eccv/RhinehartKV18}, DESIRE* \cite{DBLP:conf/cvpr/LeeCVCTC17}, MFP \cite{DBLP:conf/nips/TangS19}, MATFD \cite{DBLP:conf/cvpr/ZhaoXMCBZ0W19}, MATFG* \cite{DBLP:conf/cvpr/ZhaoXMCBZ0W19}, S-STGCNN \cite{DBLP:conf/cvpr/MohamedQEC20}, Trajectron ++ \cite{DBLP:conf/eccv/SalzmannICP20}, StarNet \cite{DBLP:conf/iros/ZhuQRX19}, TPNet \cite{DBLP:conf/cvpr/FangJSZ20}, DATF \cite{DBLP:conf/eccv/ParkLSBKFJLM20}, VectorNet \cite{DBLP:conf/cvpr/GaoSZSALS20}, CoverNet \cite{phan2020covernet}, MultiPath \cite{DBLP:conf/corl/ChaiSBA19}, NLNI \cite{DBLP:conf/iccv/Zheng0ZTN0021}, ALAN \cite{DBLP:conf/cvpr/NarayananMPLC21}, AgentFormer  \cite{DBLP:conf/iccv/0007WOK21}, and PIP \cite{DBLP:journals/corr/abs-2212-02181}. The detailed configuration of each method is demonstrated in Table. \ref{tab4} for clear presentation. Similarly, mADE and mFDE metrics are utilized for nuScenes and Argoverse datasets with 20 times of prediction. 

\begin{table}[!t]\footnotesize
  \centering
  \caption{Performance comparison on the testing set of nuScenes for @6s prediction.}
\begin{tabular}{c|c|c|c|c}
\toprule[0.8pt]

Baselines        & mADE$_5$$\downarrow$ &mADE$_{10}$$\downarrow$ &MR$_{2,5}$$\downarrow$ &MR$_{2,10}$$\downarrow$ \\
 \hline
LaPred \cite{DBLP:conf/cvpr/KimPLKKKKC21} $_{\scriptsize{\emph{CVPR2021}}}$& 1.468&1.124&0.53& 0.46\\
SGNet \cite{DBLP:journals/ral/WangWXC22} $_{\scriptsize{\emph{IEEE-RAL2022}}}$&1.863&1.399&0.67& 0.51\\
GOHOME \cite{DBLP:conf/icra/GillesSTSM22}  $_{\scriptsize{\emph{ICRA2022}}}$& 1.422 & 1.151&0.57& 0.47\\
Autobot \cite{DBLP:conf/iclr/GirgisGCWDKHP22}  $_{\scriptsize{\emph{ICRA2022}}}$& \textbf{1.368} & \textbf{1.026}&0.62&0.44\\
\hline
HRG (ours) &1.549 &1.277&0.57& 0.46\\
HRG+HSG-res. (ours) &1.535 & 1.252&0.53& 0.44\\
HRG+HSG-dot. (ours) &1.536 &1.260&\textbf{0.51}& \textbf{0.43}\\
\toprule[0.8pt]
  \end{tabular}
  \label{tab8}
  \end{table}
From the demonstrated results, the proposed method,  NLNI, and DESIRE* generate the top three performances in overall evaluation on nuScenes and Argoverse datasets. It can be observed that the comparison methods can be categorized into two groups: with or without road scene information, and most of the works encode the road scene context by the convolution network on the raw RGB images. In the meantime, from the main model architecture, besides CNN modules, RNN-based, graph-based, and attention-embedded approaches are three kinds of backbones. Among them, Graph Convolution Network (GCN) provides a promising role in trajectory forecasting, which indicates that the interaction still is the key clue for heterogeneous trajectory forecasting in mixed road scenes, and NLNI presents the best performance on the mADE evaluation (0.792 meters) because of the unlimited range consideration of interaction for road agents. Besides, because there is only the vehicle agent in the Argoverse dataset, the orientation offset in the future trajectories may be small, which can be better modeled by the Gaussian Mixture Model in the NLNI method than our Bivariate Gaussian Probability Distribution (BGPD) in future trajectory prediction. In addition, road scene context is useful for the mFDE and our method shows better mFDE than NLNI. Notably, only the road centerlines seem not enough for mixed traffic scenes owning to the different demands of various road agents, such as pedestrians and riders. Therefore, the works only with road centerlines (AgentFormer, ALAN, and CoverNet) generally show relatively poor performance compared with the ones embedded whole road scene context, such as DESIRE* and MATFG*.  For our HRG+HSG-dot. version, we consider a more fine-grained road scene context (\emph{i.e.}, road semantic graph), we outperform DESIRE*  (the third best) with 9.18\% and 25.8\% of average improvement for mADE and mFDE evaluation, respectively. Besides, owning the meaningful risk consideration, we outperform NLNI (the second best) with 3.37\% and 12.03\% of average improvement for mADE and mFDE evaluation, respectively. 

In addition, we also conduct the experiments of 6s prediction with 2s observation and compare the performance with the state-of-the-art methods in the leaderboard of the prediction task of nuScenes. In this evaluation, the methods of LaPred \cite{DBLP:conf/cvpr/KimPLKKKKC21}, SGNet \cite{DBLP:journals/ral/WangWXC22}, GOHOME \cite{DBLP:conf/icra/GillesSTSM22} , and Autobot \cite{DBLP:conf/iclr/GirgisGCWDKHP22} are compared. Table. \ref{tab8} presents the results, and we can see that our method shows comparable performance with SOTA methods and the HRG+HSG-dot obtains the best MR$_{2,5}$ and MR$_{2,10}$ value than other methods. Autobot also uses the Bivariate Gaussian Probability Distribution (BGPD) to optimize the model, while it models a latent variable sequential set encoding and decoding framework, and the decoding is fulfilled by learning the seed parameters in the discrete latent variables in Autobot. This formulation can weaken the noise influence of future prediction. 

\textbf{Effectiveness of HRG in the ApolloScape Dataset:} Besides the evaluation on nuScenes and Argoverse datasets, we also check the performance on the ApolloScape dataset. It is worth noting that there is no high-accurate map or whole scene semantic annotation in the ApolloScape dataset, the Hierarchical Scene Graph (HSG) cannot be constructed. Therefore, we compare the HRG version of the proposed method with Social-LSTM   \cite{DBLP:conf/cvpr/AlahiGRRLS16}, StarNet \cite{DBLP:conf/iros/ZhuQRX19}, S-STGCNN \cite{DBLP:conf/cvpr/MohamedQEC20}, TraPHic \cite{DBLP:conf/cvpr/ChandraBBM19}, TPNet \cite{DBLP:conf/cvpr/FangJSZ20}, GRIP++ \cite{li2019grip}, NLNI \cite{DBLP:conf/iccv/Zheng0ZTN0021}, AL-TP \cite{zhang2022ai}, and D2-TPred \cite{DBLP:conf/eccv/ZhangWGLXCM22}. Notably, the risk metrics (evaluated in Sec. \ref{rmd}) in the HRG version here omitted the Occupied Semantic Relation (OSR). We adopt wADE and wFDE as the metrics, and the comparison results are shown in Table. \ref{tab5}.
 
  \begin{table}[!t]\small
  \centering
  \caption{Performance comparison on the testing set of the ApolloScape dataset for @3s prediction.}
\begin{tabular}{c|c|c}
\toprule[0.8pt]
 \multirow{2}[4]{*}{Baselines} & \multicolumn{2}{c}{ApolloScape} \\
\cmidrule{2-3}          & wADE$\downarrow$ &wFDE$\downarrow$  \\
 \hline
Social-LSTM   \cite{DBLP:conf/cvpr/AlahiGRRLS16} $_{\scriptsize{\emph{CVPR2016}}}$ &1.890& 3.400\\
StarNet \cite{DBLP:conf/iros/ZhuQRX19} $_{\scriptsize{\emph{IROS2019}}}$&1.343 & 2.498\\
S-STGCNN \cite{DBLP:conf/cvpr/MohamedQEC20} $_{\scriptsize{\emph{CVPR2020}}}$& 1.305 & 2.344\\
\hline
TraPHic \cite{DBLP:conf/cvpr/ChandraBBM19} $_{\scriptsize{\emph{CVPR2019}}}$& 1.280 & 11.67\\
TPNet \cite{DBLP:conf/cvpr/FangJSZ20} $_{\scriptsize{\emph{CVPR2019}}}$&1.281 & 1.910\\
GRIP++ \cite{li2019grip} $_{\scriptsize{\emph{arxiv2020}}}$& 1.250 & 2.340\\
NLNI \cite{DBLP:conf/iccv/Zheng0ZTN0021}  $_{\scriptsize{\emph{ICCV2021}}}$& 1.094 & 1.545\\
AL-TP \cite{zhang2022ai} $_{\scriptsize{\emph{IEEE-TIV2022}}}$& 1.160 & 2.130\\
D2-TPred \cite{DBLP:conf/eccv/ZhangWGLXCM22} $_{\scriptsize{\emph{ECCV2022}}}$& \textbf{1.020} & 1.690\\
HRG (ours) &1.036 & \textbf{1.321}\\
\toprule[0.8pt]
  \end{tabular}
  \label{tab5}
  \end{table}
From the results, it can be observed that HRG with risk consideration outperforms other ones, especially for the wFDE evaluation, which indicates that our HRG version has a better ability for long-term prediction. It is worth noting that TrapHic, TPNet, NLNI, and our HRG all consider the group-wise moving pattern of road agents, which can leverage the similar moving pattern of road agents to avoid the isolated prediction of a single agent. Because of the behavior dependency modeling and the spatial dynamic graph modeling for each agent in D2-TPred \cite{DBLP:conf/eccv/ZhangWGLXCM22}, it can tackle the discontinuous problem of acceleration, deceleration, and turning direction, which can obtain more accurate displacement change for trajectory prediction. Therefore, D2-TPred shows a little better than our HRG in wADE metric. Among the comparison, NLNI and our HRG version consider the between-category moving pattern relation, which implies safer trajectory forecasting and better wFDE values are obtained.

\begin{table*}[!t]\scriptsize
  \centering
  \caption{Performance comparison, w.r.t., road agent category, on the testing set of the nuScenes dataset. }
\renewcommand{\arraystretch}{1.2}
\setlength{\tabcolsep}{1.5mm}{
\begin{tabular}{c|c|c|c||c|c|c}
\toprule[0.8pt]
 \multirow{2}[4]{*}{Baselines} & cars-3s &pedestrians-3s & riders-3s& cars-6s &pedestrians-6s & riders-6s\\
\cmidrule{2-7}          & mADE/mFDE/MR$_{2,20}$& mADE/mFDE/MR$_{2,20}$ & mADE/mFDE/MR$_{2,20}$& mADE/mFDE/MR$_{2,20}$& mADE/mFDE/MR$_{2,20}$ & mADE/mFDE/MR$_{2,20}$  \\
 \hline
S-STGCNN \cite{DBLP:conf/cvpr/MohamedQEC20} &1.35/2.11/0.60&0.48/\textbf{0.40}/0.14&2.21/3.73/0.45&1.65/2.51/0.67&0.58/0.70/0.16&2.54/3.82/0.60\\
\hline
HRG &1.29/1.91/0.35&\textbf{0.44}/0.45/\textbf{0.06}&1.16/1.58/0.28&1.29/1.92/0.52&\textbf{0.46}/\textbf{0.46}/\textbf{0.07}&1.34/1.67/0.32\\
HRG+HSG-res.&1.28/1.89/\textbf{0.34}&0.48/0.44/0.06&0.81/1.22/\textbf{0.26}&\textbf{1.25}/\textbf{1.81}/\textbf{0.50}&0.48/0.51/0.08&0.80/1.12/0.34\\
HRG+HSG-dot.&\textbf{1.11}/\textbf{1.62}/0.34&0.47/0.42/0.07&\textbf{0.72}/\textbf{1.00}/0.27&1.26/1.89/0.52&0.48/0.53/0.11&\textbf{0.79}/\textbf{1.05}/\textbf{0.29}\\
\toprule[0.8pt]
  \end{tabular}}
  \label{tab6}
  \end{table*}
 \begin{figure*}[!t]
  \centering
 \includegraphics[width=\hsize]{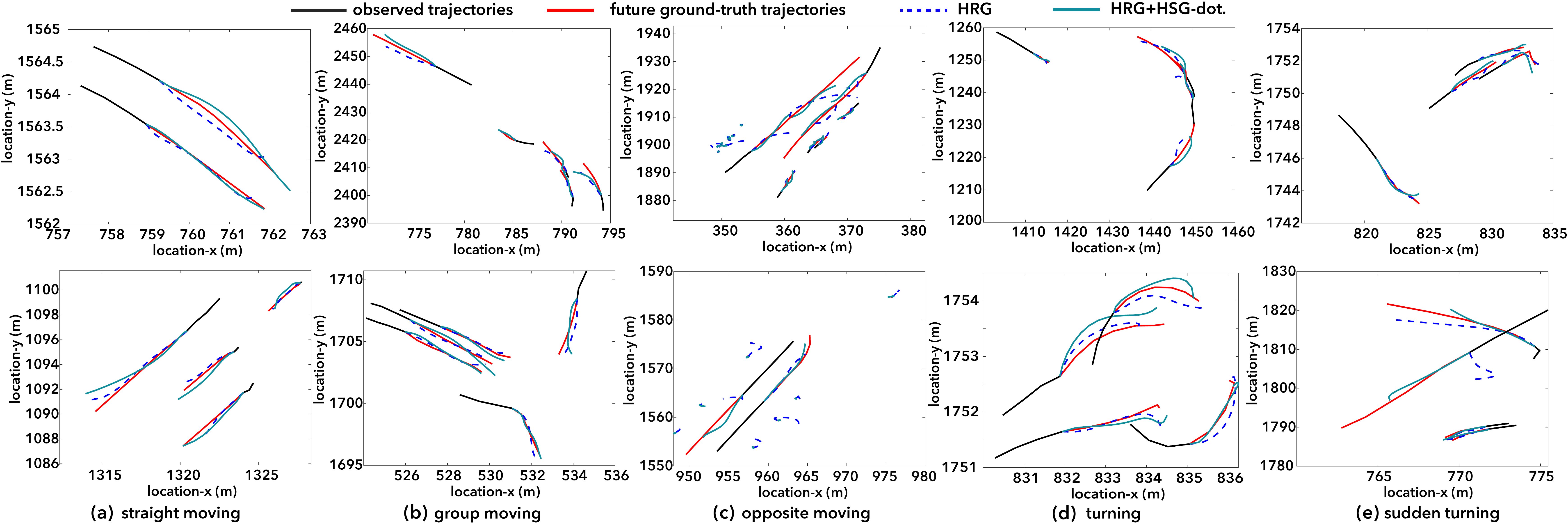}
  \caption{Some typical samples of forecasted trajectories (@3s) in nuScenes dataset with the minimum ADE (20 predictions) by HRG and HRG+HSG-dot.}
  \label{fig10}
\end{figure*}

\subsection{Evaluation on Heterogeneous Agents}
Except for the overall performance evaluation, we also present the evaluation of different versions of the proposed method. To check the performance for heterogeneous agents, the classic S-STGCNN \cite{DBLP:conf/cvpr/MohamedQEC20} is used for a pure comparison with HRG and HSG. In addition, we provide two kinds of cases of prediction, \emph{i.e.}, 3s-prediction and 6s-prediction with the same length of 2s-observation. 

The results are shown in Table. \ref{tab6}. From this table, we find that for short horizon prediction, the scene graph (playing more role in \emph{dot.} fusion than \emph{res.} fusion) can assist the prediction for 3s case, while the risk graph (with \emph{res.} fusion) helps the prediction of pedestrians and cars more. For pedestrians, we can see that the MR values change little for the 3s and 6s predictions because of the small displacement. Besides, different versions of the proposed method show similar performance for pedestrians because of the little scene change in short displacement. Therefore, the heterogeneous risk graph may take the core role for pedestrians. As for cars, MR values for the 6s prediction become larger than the 3s case owning to the fast velocity and uncertain behaviors. For riders, because of the flexible movement regions (e.g., frequent switch between the sidewalk and the road) and fast velocity, the scene graph may change frequently and provide semantic layout guidance. Therefore,  ``HRG+HSG-dot." shows the best for riders. However, because of the higher uncertainty for longer-term prediction, riders obtain larger MR values.  

We also show some snapshots of forecasted trajectories by the proposed method, as shown in Fig. \ref{fig10}. For a clear comparison, we compare HRG and HRG+HSG-dot. (with the most consideration of HSG) here. Because of the sampling way of Bivariate Gaussian Probability Distribution, the predicted trajectories may have some small fluctuations, which will weaken the future decision accuracy of cars. Although some works can take the physical kinematic constraint to reduce this phenomenon, the fluctuations are hard to be avoided by the trajectory distribution sampling process. This work introduces the \textbf{Bezier Curve} \cite{bezier2014mathematical} to reduce the fluctuations with 6-order degrees (six points in each trajectory) as
\begin{equation}
B(\sigma)=\sum_{t=0}^n(^n_t){\hat{\bf{p}}}_t(1-\sigma)^{n-t}\sigma^t, t\in[1,6], \sigma\in[0,1],
\end{equation}
where $n$ denotes the orders of the Bezier Curve (set as 6 in this work) and $\sigma$ is the updating variable in Bezier Curve. From this figure, the generated trajectories from ``HRG+HSG-dot." are more physically traversable from the presented five behavior categories of \emph{straight moving}, \emph{group moving}, \emph{opposite moving}, \emph{turning}, and \emph{sudden turning}.

\section{Conclusions}
\label{con}
Aiming at heterogeneous trajectory forecasting in driving scenarios, this paper modeled a Heterogeneous Risk Graph (HRG) and a Hierarchical Scene Graph (HSG) to infer the road constraint and social relation of road agents with the environment. HRG is modeled to explore the risk clue of moving direction, distance, occupied semantic regions, and moving pattern relation of different road agents. HSG is modeled by the alignment of the road agents and road semantic regions to the road scene grammar. With the HRG and HSG encoding by spatial-temporal graph convolution network, their embeddings are fused and decoded for future trajectory forecasting.  By this formulation, we exhaustively evaluated the performance of the proposed method with other state-of-the-art, and the superiority and the role of HSG are manifestly verified. 

Actually, for the heterogeneous trajectory forecasting in mixed traffic scenes, the movement of different kinds of road participants depends on their moving intention. Therefore, in the future, intention-driven heterogeneous trajectory forecasting will be investigated. In addition, trajectory forecasting has a practical promotion for safe driving, we also will investigate the promising models for traffic accident prediction involving trajectory forecasting, as well as their counterfactual analysis.

\balance
{\small
\bibliographystyle{IEEEtran}
\bibliography{ref}
}

\end{document}